\documentclass[default,iicol,sn-chicago]{sn-jnl}

\usepackage{comment}
\usepackage{subfigure}
\usepackage{commath}
\usepackage{amsmath}
\usepackage{multirow}
\usepackage{graphicx}
\usepackage{times}
\usepackage{epsfig}
\usepackage{graphicx}
\usepackage{amsmath}
\usepackage{amssymb}
\usepackage{pifont}
\usepackage[rightcaption]{sidecap}
\usepackage{verbatim}
\usepackage{paralist}
\usepackage{array,longtable}
\usepackage{colortbl}
\definecolor{applegreen}{rgb}{0.62, 0.82, 0.0}
\definecolor{nonerikblue}{rgb}{0.33, 0.36, 0.87}
\jyear{2021}%

\theoremstyle{thmstyleone}%
\theoremstyle{thmstyletwo}%

\theoremstyle{thmstylethree}%

\begin{document}

\title[Article Title]{The Right Spin: Learning Object Motion from Rotation-Compensated Flow Fields}

%Univ. Grenoble Alpes, Inria, CNRS, Grenoble INP, LJK, 38000 Grenoble, France.
%Inria, École normale supérieure, CNRS, PSL Research University,  Paris, France
%University of Massachusetts Amherst, USA 

\author*[1]{\fnm{Pia} \sur{Bideau}}\email{p.bideau@tu-berlin.de}

\author[2]{\fnm{Erik} \sur{Learned-Miller}}

\author[3]{\fnm{Cordelia} \sur{Schmid}}

\author[4]{\fnm{Karteek} \sur{Alahari}}

\affil*[1]{\orgdiv{Science of Intelligence}, \orgname{TU Berlin}, \orgaddress{\street{Marchstr}, \city{Berlin}, \postcode{10587}, \country{Germany}}}

\affil[2]{\orgname{University of Massachusetts}, \orgaddress{\street{Governors Dr}, \city{Amherst}, \postcode{01002}, \state{MA}, \country{US}}}

\affil[3]{\orgname{Inria, {\'E}cole Normale Sup\'erieure, CNRS, PSL Research University}, \orgaddress{\city{Paris}, \country{France}}}

\affil[4]{\orgname{Univ. Grenoble Alpes, Inria, CNRS, Grenoble INP, LJK}, \orgaddress{\postcode{38000}, \state{Grenoble}, \country{France}}}

%\affil[3]{\orgdiv{Department}, \orgname{Organization}, \orgaddress{\street{Street}, \city{City}, \postcode{610101}, \state{State}, \country{Country}}}


\abstract{Both a good understanding of geometrical concepts and a broad familiarity with objects lead to our excellent perception of moving objects. The human ability to detect and segment moving objects works in the presence of multiple objects, complex background geometry, motion of the observer and even camouflage. How humans perceive moving objects so reliably is a longstanding research question in computer vision and borrows findings from related areas such as psychology, cognitive science and physics.
One approach to the problem is to teach a deep network to model all of these effects. This contrasts with the strategy used by human vision, where cognitive processes and body design are tightly coupled and each is responsible for certain aspects of correctly identifying moving objects.
Similarly from the computer vision perspective, there is evidence~\citep{bideauCVPR2018,irani1998unified} that classical, geometry-based techniques are better suited to the ``motion-based'' parts of the problem, while deep networks are more suitable for modeling appearance. In this work, we argue that the coupling of camera rotation and camera translation can create complex motion fields that are difficult for a deep network to untangle directly. 
We present a novel probabilistic model to estimate the camera's rotation given the motion field. We then rectify the flow field to obtain a rotation-compensated motion field for subsequent segmentation. This strategy of first estimating camera motion, and then allowing a network to learn the remaining parts of the problem, yields improved results on the widely used DAVIS benchmark as well as the recently published motion segmentation data set MoCA (Moving Camouflaged Animals).}

\keywords{Motion segmentation, Video segmentation, Optical flow, Camera motion estimation}



\maketitle

% ---------------------------------------------------------------
\section{Introduction}
\begin{comment}
\begin{figure}
\begin{center}
    \subfigure[flow]
    {
        \includegraphics[width=0.3\columnwidth]{fig/ICCV19/flow_goat_00017.png}
    }
    \subfigure[\scriptsize rotation-comp. flow]
    {
        \includegraphics[width=0.3\columnwidth]{fig/ICCV19/flowTrans_goat_00017.png}
    }
    \subfigure[frame]
    {
        \includegraphics[width=0.3\columnwidth]{fig/ICCV19/00016.jpg}
    }
    \subfigure[angle of (b)]
    {
        \includegraphics[width=0.3\columnwidth]{fig/ICCV19/angle_goat_00017.png}
    }
    \subfigure[angle of (b)]
    {
        \includegraphics[width=0.3\columnwidth]{fig/ICCV19/angleTrans_goat_00017.png}
    }
    \subfigure[segmentation]
    {
        \includegraphics[width=0.3\columnwidth]{fig/ICCV19/goat-segmentation-30.png}
    }
   \caption{{\bf What is moving?} Coupling of camera rotation and camera translation often create complex motion fields that are difficult for a network to untangle. Instead we propose a strategy to learn object motion patterns based on rotation compensated flow.}
   \label{fig:algo_overview}
\end{center}
\end{figure}
\end{comment}

\begin{figure}
\begin{center}
    \subfigure[flow]
    {
        \includegraphics[width=0.29\columnwidth]{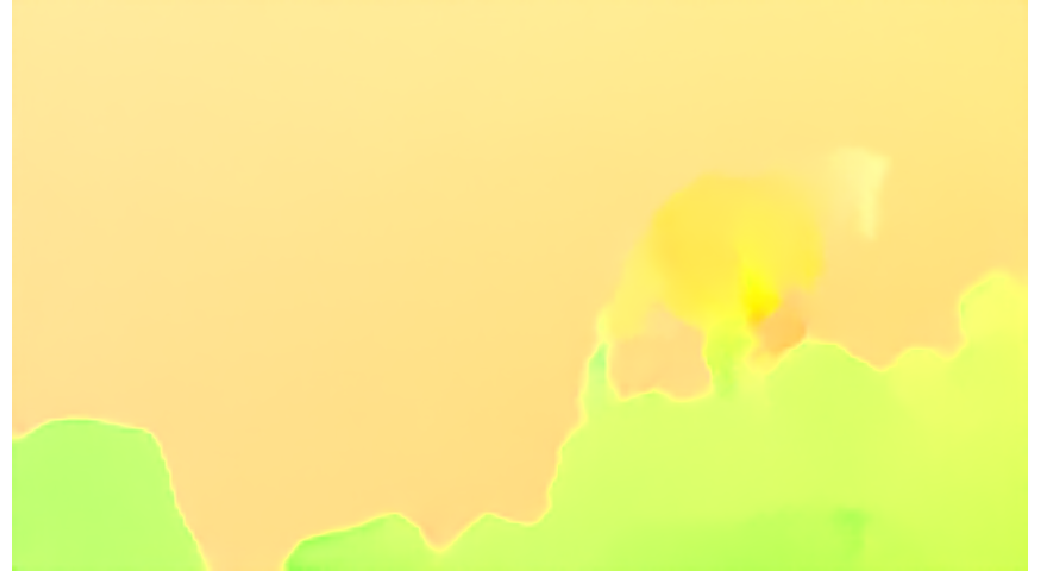}
    }
    \subfigure[rot.-comp. flow]
    {
        \includegraphics[width=0.29\columnwidth]{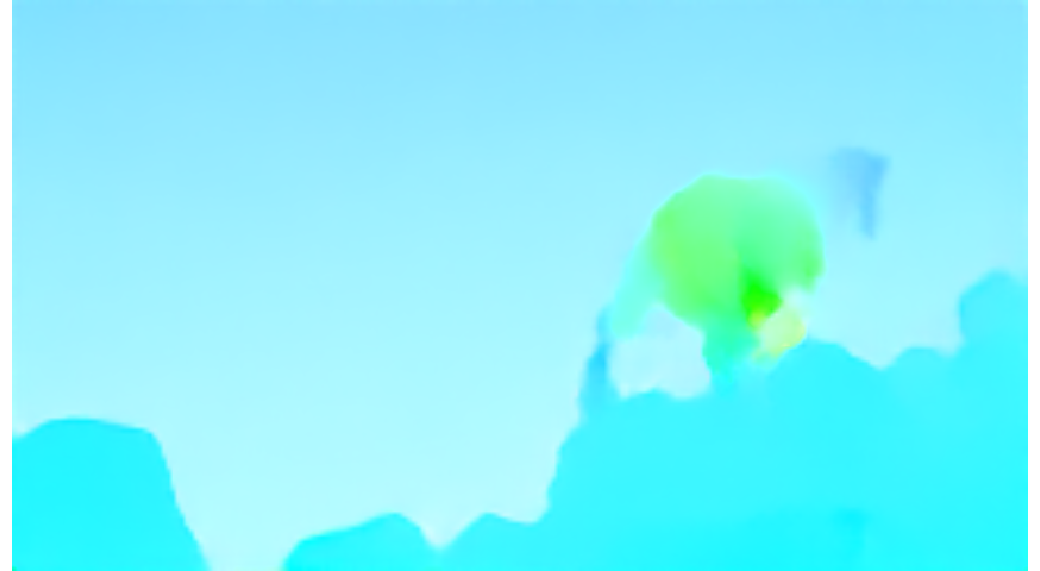}
    }
    \subfigure[frame]
    {
        \includegraphics[width=0.29\columnwidth]{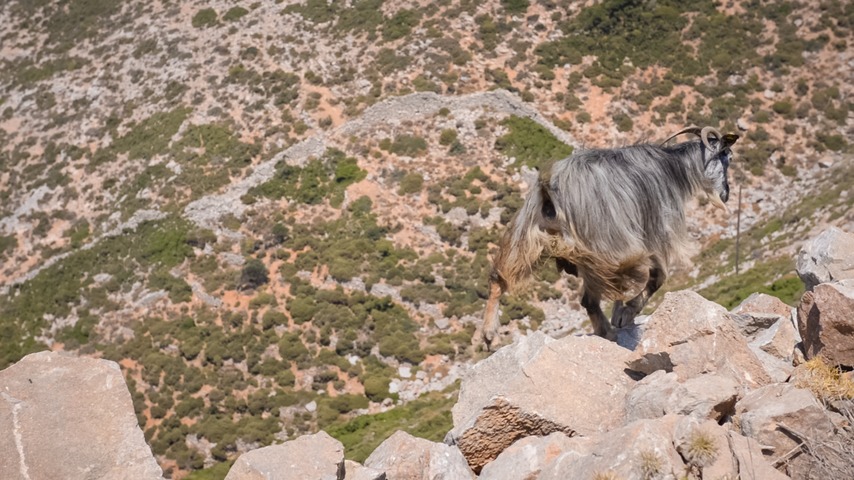}
    }
    \subfigure[angle of (a)]
    {
        \includegraphics[width=0.29\columnwidth]{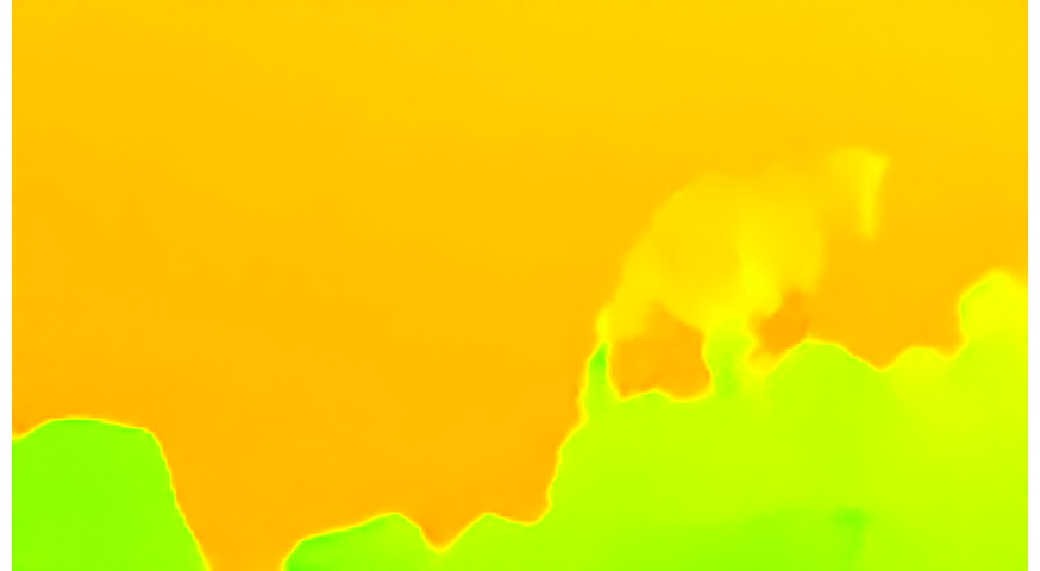}
    }
    \subfigure[angle of (b)]
    {
        \includegraphics[width=0.29\columnwidth]{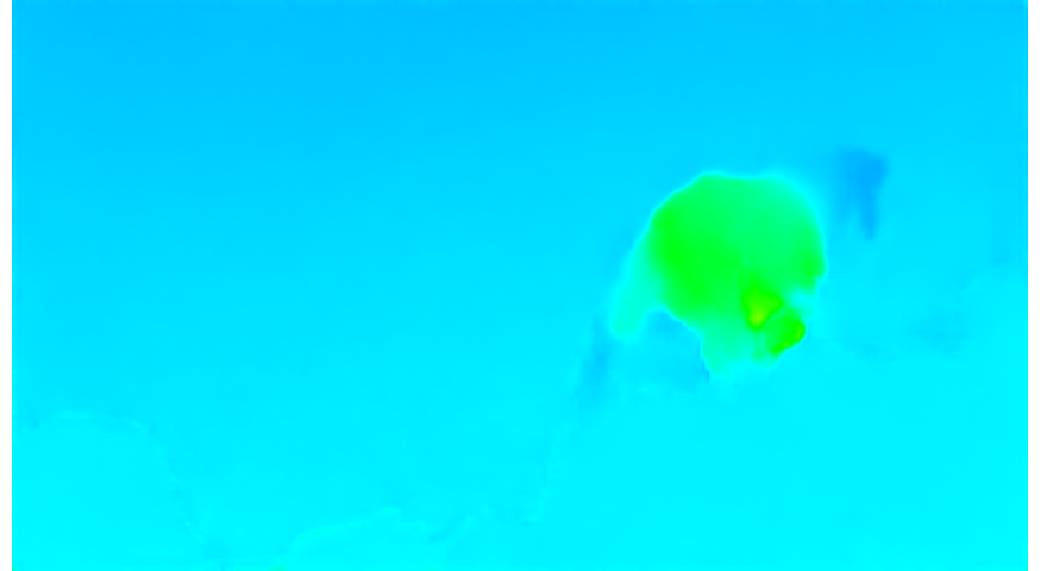}
    }
    \subfigure[segmentation]
    {
        \includegraphics[width=0.29\columnwidth]{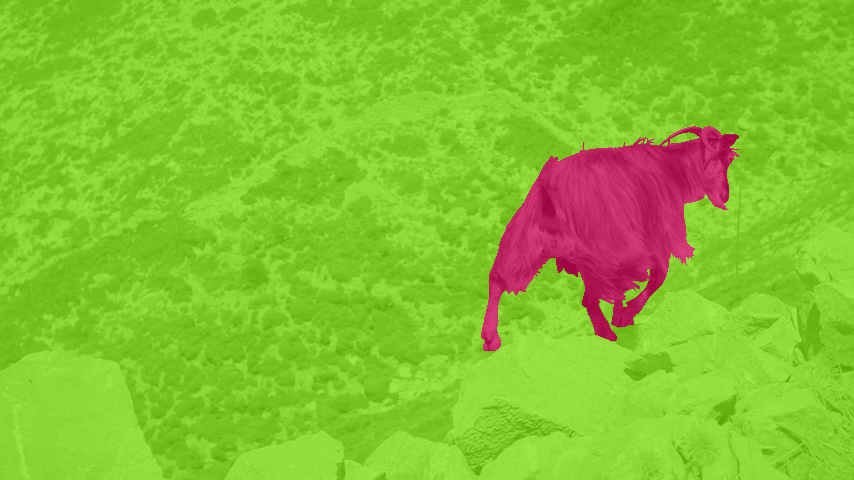}
    }
   \caption{{\bf What is moving?} Coupling of camera rotation and camera translation often create complex motion fields that are difficult for a network to untangle. Instead we propose a strategy to learn object motion patterns based on rotation compensated flow.}
   \label{fig:algo_overview}
\end{center}
\end{figure}

The human visual system has the ability to detect independently moving objects within a high variety of different environments. 
While we are moving through the world our eye captures a large amount of visual information over time. 
Often, we are not aware of the remarkable preprocessing steps that happen almost unnoticed. 
For example, human eye movements induce two major simplifications to incoming images before visual information is processed by the visual cortex. These are (1) stabilizing the image - reducing the amount of local change due to motion, and (2) changing the direction of gaze \citep{walls1962evolutionary, longuet1980interpretation}. 

Here, we revisit this approach to motion segmentation that separates the problem into two parts: first, we preprocess the perceived motion field following well known geometrical concepts leading to important simplifications and second, learning to segment independently moving objects from these simplified motion fields. %\textcolor{red}{Erik: The previous sentence makes it sound like this is the first paper to do motion compensation. Perhaps it should be rephrased? You could say something like 'Here, we revisit the approach to motion segmentation that separates the problem into two parts.... We propose several novel improvements to these methods that lead to significant improvements.' Or something like this.}

{\it In computer vision, the task of motion segmentation attempts to analyze the perceived motion and to segment a video sequence into static environment (if any) and independently moving objects} \citep{bideau2016detailed}. Interpreting the motion field accurately, and then drawing the right conclusions about what is moving in the world and what is static, is a complex process. Even in biological vision systems applied strategies are still only partially understood.

Unlike most end-to-end learning-based approaches, where a model learns all necessary steps between the input and the final output, we break down the problem of motion segmentation into two sub-problems: adjusting the optical flow to remove the effects of camera rotation ({\em rotation compensation}) using classical approaches based on perspective projection and learning to segment the remaining optical flow into static background and moving objects. The step of compensating
for camera rotation is a challenging one, since the flow field is only a noisy estimate of the motion field \citep{bideauCVPR2018,bideau2016s}. In cases of little motion or of featureless areas, the observed flow field is often erroneous and thus the true camera motion and object motion is hard to estimate accurately. 

To this end, we present a novel probabilistic method for estimating camera rotation and derive a new likelihood function modeling the probability of an observed optical flow field, given our estimated (ideal) motion field. A CNN framework is then integrated for learning to segment moving objects after the motion of the camera has been determined. 

Our contributions include: (i)~estimating the camera rotation and translational motion direction in the presence of moving objects, using a new likelihood maximization approach, (ii)~given the rotation compensated flow, we show that the task of learning motion patterns is improved, resulting in a better motion segmentation performance shown on two data sets - the widely used DAVIS benchmark~\citep{perazzi2016benchmark} and the recently published data set MoCA~\citep{lamdouar2020betrayed}.

The paper is structured as follows. In Section~\ref{sec:relatedwork} we review relevant work on motion segmentation starting from classical geometry based approaches and concluding with the most recent work using convolutional neural networks to segment moving objects from optical flow. 
In Section~\ref{sec:ICCV19}, we develop an end-to-end approach for motion segmentation.
We briefly review the basics about the motion field and how it is related to camera motion, depth and object motion (Section~\ref{sec:motionfield}). Building upon key concepts of perspective projection the methodological approach is derived in two subsections:
estimating the camera rotation to produce rotation-compensated flow fields (Section~\ref{sec:cam-estimation}) and segmenting the remaining (noisy) translational flow field into independently moving objects and static background (Section~\ref{sec:learning-motion}).
A multifaceted evaluation of the proposed approach, including multiple ablation studies has been carried out and is shown in Experiments (Section~\ref{sec:experiments}).

\begin{figure*}
\begin{center}
        \includegraphics[width=0.9\textwidth]{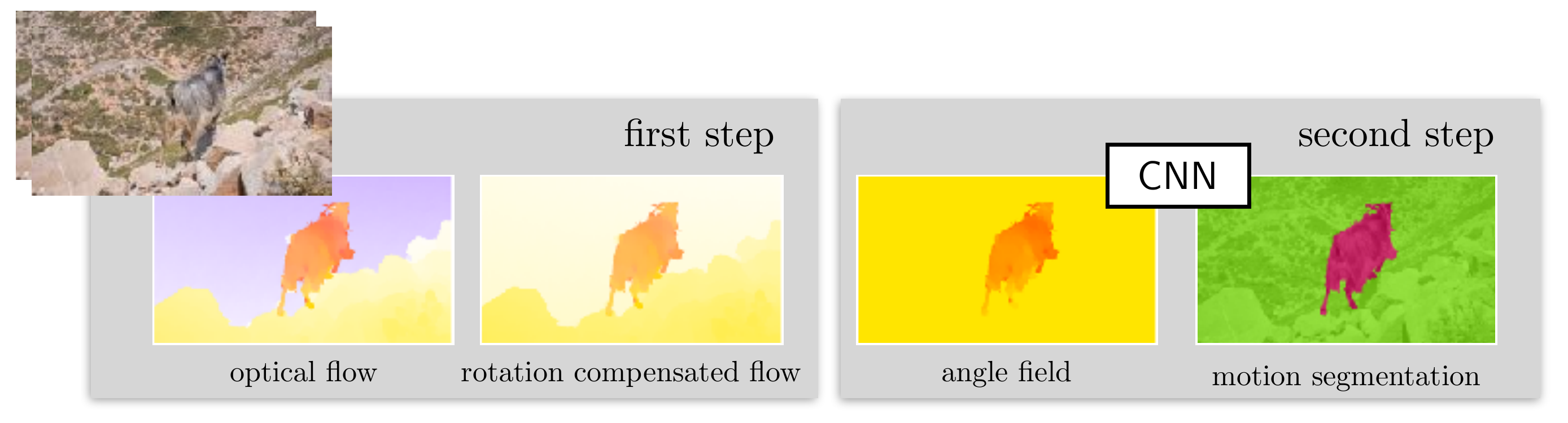}
   \caption{{\bf Getting the right spin.} We first compensate the observed motion field for camera rotation (``first step''), and segment the remaining translational optical flow field using a learning based approach (``second step''). The observed flow field on the  left has complex motion patterns: the motion directions of foreground and background are pointing in opposite directions, due to large variance in scene depth, and the combined impact of camera rotation and translation. Estimating the camera rotation (``the right spin''), and compensating the flow field for this rotation simplifies the motion field dramatically, in this case yielding similar motion directions for foreground and background. This provides simpler inputs to our learning based motion segmentation framework.}
   \label{fig:overview}
\end{center}
\end{figure*}

% ---------------------------------------------------------------
\section{Related Work}\label{sec:relatedwork}
Many works tackling the problem of motion segmentation focus on \textit{binary motion segmentation}, where pixels are classified as either moving or being part of the static background. In that case no distinction is made between differently moving objects~\citep{bideau2016s,narayana2013coherent,papazoglou2013fast,faktor2014video}. Others~\citep{taylor2015causal,keuper2015motion,fragkiadaki2012video} address {\em multi-label motion segmentation}, where a separate label is given to each independently moving object. Our work tackles binary motion segmentation, but we consider both views onto the segmentation problem in this review of related work.

\subsection{Classical approaches}

\paragraph{Methods based on feature clustering}\label{ref:sec-literature-trajectory} 
To capture motion information, typically point trajectories are either formed by tracked image features or dense optical flow. 
Then trajectories sharing similar motion characteristics are grouped into coherent motion clusters describing the motion of a particular object~\citep{keuper2015motion,brox2010object,fragkiadaki2012video,ochs2011object,keuper2017,yan2006general,shen2018submodular,lezama2011track}. 

These approaches vary in defining typical motion characteristics for clustering. Yan et. al \citep{yan2006general} propose to cluster trajectories based on geometric constraints (trajectories of the same motion lie in a manifold) and locality. In \citep{keuper2017,keuper2015motion} the segmentation problem is represented as a minimum cost multicut graph problem, where edge weights are computed from motion, position and color cues.

These trajectory based clustering approaches reach their limit if understanding of the scene structure is necessary to segment a moving object correctly. Trajectories perfectly represent long-term pixel displacements between a sequence of frames.
Pixel displacements however are a function of depth and motion. Thus trajectory based clustering methods often form clusters not only for independently moving objects, but also for objects at different depths. For instance if the camera is translating and rotating rocks close to the camera produce a very different flow pattern that the far away scene (see Figure \ref{fig:overview}), thus those two areas would form two separate clusters although neither the rock nor the far away scene is moving.

Methods based on occlusions~\citep{ogale2005motion,taylor2015causal} are subject to similar depth-related problems, since occlusions could be caused at depth boundaries as well as motion boundaries. A distinction is often not made.

\paragraph{Methods based on projective geometry} \label{ref:sec-literature-projective}
Projective geometry is an extension of the Euclidean and affine space and contains properties of perspective projection. 
It is widely used as a mathematical formalism  to describe the geometry of cameras and its associated transformations \citep{torr1998geometric,zamalieva2014background,wang1994representing,ke2002robust,jin2008background,xiao2005motion,vidal2004unified,xu2018motion}. 

Different from trajectory based clustering methods, motion segmentation approaches relying on projective geometry analyze the optical flow
between a pair of frames, grouping pixels into regions where flow is
consistent with motion models that are explainable by projective geometry \citep{torr1998geometric,zamalieva2014background,wang1994representing,ke2002robust,jin2008background,xiao2005motion,xu2018motion}.
Torr~\citep{torr1998geometric} develops a sophisticated probabilistic
model of optical flow, building a mixture model
that explains an arbitrary number of rigid components within the
scene. Interestingly, he assigns different types of motion models to
each object based on model fitting criteria. 
Zamalieva et al.~\citep{zamalieva2014background} and Xun Xu et al.~\citep{xu2018motion} present a combination of methods that rely on both - projective geometry (homography estimation) and perspective projection (fundamental matrix estimation). The two methods have complimentary
strengths, and the authors attempt to select among the best
dynamically.

Methods relying on projective geometry perform well in cases of planar motion (motion obtained by a translating or rotating camera picturing a planar scene or a very distant scene, where effects of 3D parallax are negligible), however similarly to cluster based approaches these methods fall short in case of complex scene geometry.

Horn identified specific drawbacks of using projective geometry in such estimation problems and has argued that methods based directly on
perspective projection are less prone to overfitting in the presence of noise \citep{horn1999projective}.  

\paragraph{Methods based on perspective projection} \label{ref:sec-literature-perspective}
Perspective geometry allows us to mathematically explain and model the process of how the three-dimensional world is projected on to a just two-dimensional image plane. Artists and scientists like Alberti, Brunelleschi, D\"urer and da Vinici studied effects of perspective projection about 500 years back in time~\citep{pirenne1952scientific}. These insights have made a significant contribution to current successes in computer vision.
One of the key aspects of perspective projection is the observation that two parallel lines (in the euclidean space) are transformed to two lines that intersect in the vanishing point at the horizon on the image plane.

It has been shown that motion segmentation approaches based on perspective projection \citep{irani1998unified,bideau2016s,bideauCVPR2018,narayana2013coherent,vidal2002segmentation,zhang2007moving, yang2021learning} are more accurate (in terms of model agreement to the physical world) than those based on projective geometry, since the latter omits certain constraints in modeling image transformations~\citep{horn1999projective,bideau2016s}. Having a model that is confirm with the physical world might be especially critical for tasks where interaction with the physical world is required in a second step such as in robotics or autonomous driving.

\subsection{Learning motion segmentation using convolutional neural networks}

\paragraph{Methods based on supervised learning}
Several approaches as~\citep{Tokmakov17,Tokmakov17ICCV,fusionseg,Cheng_ICCV_2017,dave2019towards,ranjan2019competitive,vertens2017smsnet, Mahadevan20BMVC, lamdouar2020betrayed, cheng2017segflow} 
have explored the strength of deep neural networks to learn motion patterns of moving objects and to produce binary motion masks distinguishing whether a pixel belongs to a moving object or not. Most approaches propose a two-stream architecture~\citep{Tokmakov17ICCV,fusionseg,dave2019towards} to separately process motion and appearance.

\noindent Theses approaches learn motion patterns given the optical flow, the raw video frames or optical flow together video frames. Rather than following the true physics of image formation, convolutional neural networks are able to learn high level motion patterns of background motion and object motion. 
This ability has the clear advantage of not being dependent upon technical camera parameters such as the focal length or image distortions due to various lens characteristics or constraints induced by technical parts of the camera (mechanical or electronic). 

\paragraph{Methods based on self-supervised learning}
General concerns of deep-learning based approaches and in particular supervised approaches are overfitting to a particular type of object category that is likely to move~\citep{dave2019towards} and the lack of large amounts of training data. 
To overcome the problem of limited training data, two straight forward approaches are either using synthetic training data~\citep{Tokmakov17,Tokmakov17ICCV} or relying on noisy estimates of the motion field~\citep{fusionseg} using other algorithms~\citep{sun2018pwc,ilg2017flownet,sun2010secrets}. However, both paths are still in need of large amounts of training data (although no additional manual annotations are required in these cases), this rises the need for self-supervised approaches~\citep{yang2021self, lu2019see, yang2019unsupervised, lai2020mast,gordon2019depth,Bideau_2018_ECCV_Workshops}.
Incorporating knowledge about the real world physics into the training procedure of a neural network is an alternative to various kinds of data augmentation approaches that is subject of current research~\citep{tung2019learning, yang2021learning}. Some of those ideas have been already successfully applied in context of self-supervised learning~\citep{gordon2019depth,Bideau_2018_ECCV_Workshops}. 

Here, we propose a novel approach to the motion segmentation problem that specifically combines aspects of perspective projection and learns general object motion patterns.

% ---------------------------------------------------------------
\section{Learning object motion from rotation-compensated flow}\label{sec:ICCV19}
\begin{figure}
\begin{center}
\includegraphics[width=\columnwidth]{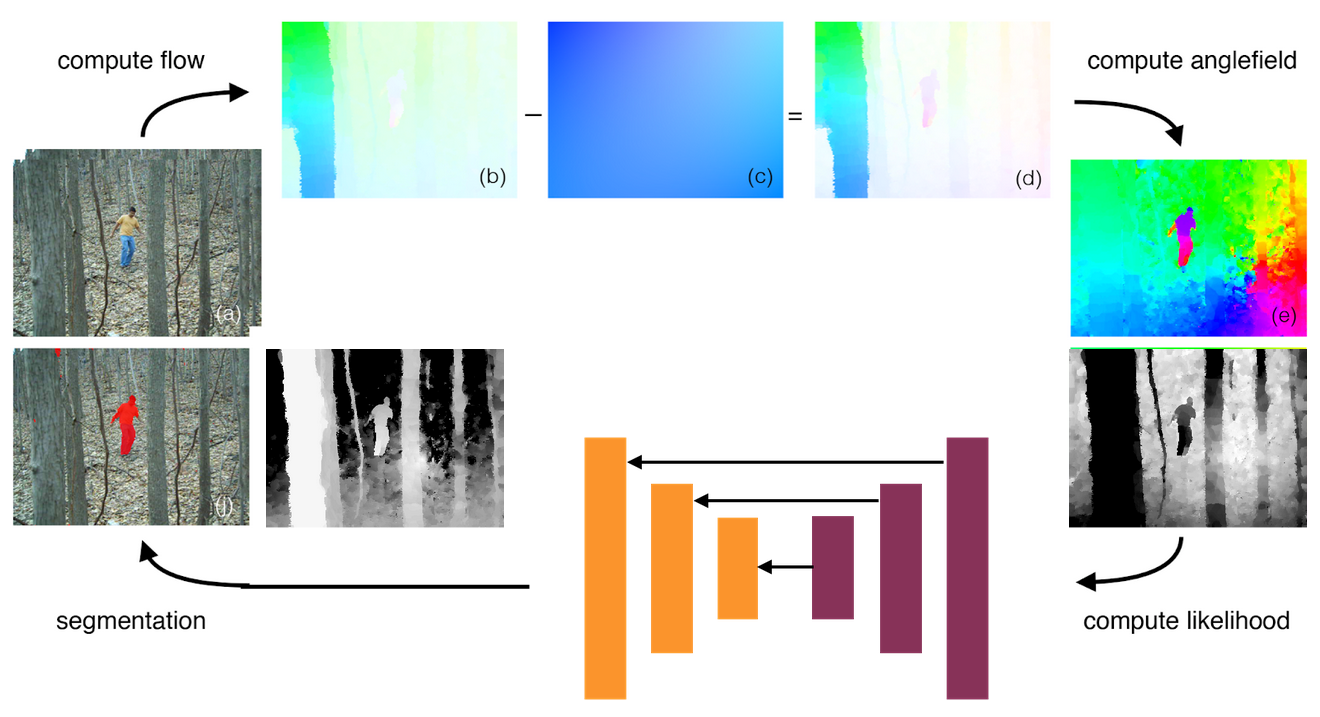}
   \caption{{\bf Overview of our approach.} Given the optical flow (b) the camera rotation is estimated (Section~\ref{sec:cam-est-trans}). The flow $\vec{v_r}$ due to camera rotation is defined by the motion parameters $(A,B,C)$. (c) is subtracted from the optical flow $\vec{o}$ to produce a translational flow $\vec{o}_t$. The flow angle $\theta_{\vec{o}_t}$ and magnitudes $\abs{\vec{o}_t}$ are shown in (e).}
   \label{fig:overview-moa}
\end{center}
\end{figure}

As most of the previous works we define a \textit{moving object} as a \textit{collection of matter that independently moves as a whole in the 3D world}.
An overview of our approach for motion segmentation is shown in Figure~\ref{fig:overview-moa}. Given an estimate of the motion field (optical flow) each frame is segmented into static environment and independently moving objects.
To achieve this we present an approach where we first estimate the 
camera rotation and then use this knowledge to form a rotation-compensated flow field. A network is trained that takes rotation-compensated flow fields as input and outputs motion segmentation masks. To this end, we combine our novel geometry-based method for estimating camera rotation, and a CNN framework for learning to segment moving objects. 

In the following we will revise relevant background information about the formation of a motion field, that occurs on the camera sensor as the camera moves (Section~\ref{sec:motionfield}). Building on this, we propose a novel approach to estimate camera rotation in complex environments, considering scene depth as well as independently moving objects (Section~\ref{sec:cam-estimation}).
In Section~\ref{sec:learning-motion}, we propose an approach similar to~\citep{Bideau_2018_ECCV_Workshops} that learns to segment the rotation compensated motion field into static background and independently moving objects.

%-----------------------------------------------------------------------
\subsection{The Motion Field: A Geometrical Analysis}\label{sec:motionfield}
The motion field captures pixel displacements between two consecutive frames. Displacements arise typically due to one of the following factors: (1)~a moving camera, (2)~one or more objects moving in the 3D world. These pixel displacements depend not only on the speed of objects or the camera, but also the scene geometry. 

As an example to illustrate the different factors that influence the formation of the motion field, let's consider the ``goat'' sequence from the DAVIS data set (Figure~\ref{fig:overview}). Based on the original flow field it is hard to estimate which pixels belong to the moving object and which belong to static background. The direction of the flow in the background region differs significantly from the flow describing the motion of the rocks in the foreground region (motion direction is color encoded). However, neither the background nor the rocks are moving differently in the 3D world. To detect objects that are actually moving independently in 3D it is necessary to decompose the observed motion field.
We formalize these observations and review the geometrical construction of the motion field. 

\subsubsection{Motion field}
Let $[U,V,W]$ be the parameters describing the camera translation and $[A,B,C]$ the parameters describing camera rotation\footnote{The rotation parameters are often referred to as pitch, yaw and roll.} along the x, y and z axes respectively. Let $f$ be the camera's focal length and $Z$ the relative scene depth at a pixel location $(x,y)$. In this setting, the motion vector $\vec{v}$ due to camera motion is given by:
\begin{align}
\vec{v} &= \vec{v}_r + \vec{v}_t = 
\begin{pmatrix}
  u_r  \\
  v_r 
 \end{pmatrix} +
 \begin{pmatrix}
  u_t  \\
  v_t 
 \end{pmatrix},\\
 &= 
 \begin{pmatrix}
  \frac{A}{f}xy  - Bf - \frac{B}{f}x^2 + Cy \\
  Af + \frac{A}{f}y^2 - \frac{B}{f}xy - Cx
 \end{pmatrix}
 +
 \begin{pmatrix}
  \frac{-fU+xW}{Z}\\
 \frac{-fV+yW}{Z}
 \end{pmatrix},
 \label{eqn:flowdecomp}
\end{align}
where $\vec{v}_r$ and $\vec{v}_t$ represent motion field vectors corresponding to camera rotation and translation respectively. Equation~(\ref{eqn:flowdecomp})\footnote{In fact, this equation is an approximation, and only holds if the rotation angles are small~\citep{longuet1980interpretation}. To obtain the exact rotational flow field one has to transform the 2D image points to 3D using perspective projection equations, rotate the points according to the camera's rotation in 3D, backproject them onto the 2D image plane, and then measure the displacement.} highlights an important properly, namely that the flow due to camera rotation is only determined by the camera rotation parameters and the camera's focal length. {\it  The flow due to camera rotation is independent of the scene depth.}  One can subtract this rotational motion component at each pixel to obtain a rotation-compensated flow field. 

\subsubsection{Rotation-compensated motion field}\label{sec:motionfield-anglefield}
As shown in the flow equation (\ref{eqn:flowdecomp}), the rotation-compensated flow field $\vec{v}_t$ is determined by the translational camera motion $[U,V,W]$, and the scene depth $Z$. It comprises all the relevant information about the scene geometry, unlike the rotational component $\vec{v}_r$, which is independent of the scene geometry. The magnitude of the rotation-compensated flow is inversely related to scene depth, i.e., regions further away from the camera have small translational flow magnitude, and those closer to the camera have larger magnitudes. {\it The direction of $\vec{v_t}$ (flow angle) however does not depend upon the scene depth}:
\begin{align}
\theta=
\begin{cases}
  \arccos{(xW - fU)}, & \text{if}\ (yW - fV)>0, \\
  2\pi-\arccos{(xW - fU)}, & \text{otherwise.}
\end{cases}
\label{eq:anglefield}
\end{align}

\begin{figure}[h]
\begin{center}
   \includegraphics[width=0.99\columnwidth]{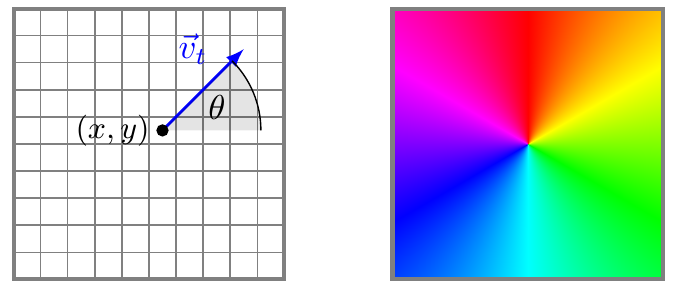}
   \caption{{\bf Translational motion field vector.} Left: motion field vector $\vec{v}_t$ at pixel position $(x,y)$. Right: color coding of the angle field $\theta(x,y)$ at each pixel location for the case of camera translation along the optical axis, i.e. $[U,V,W]=[0,0,1]$.}
   \label{fig:motion-vec-def}
\end{center}
\end{figure}

Figure \ref{fig:motion-vec-def} pictures the computation of the flow angle $\theta$ at pixel locations $(x,y)$, leading to an angle field as shown on the right. Where as Figure~\ref{fig:motion-vec-def} pictures the angle field of pure camera translation, Figure~\ref{fig:overview} shows an angle field of a scene with camera translation and object motion. Here, independently moving objects, can be observed as discontinuities in angle. The angle of the rotation-compensated flow alone is independent of the scene geometry, thus independently moving objects stand out due to their different direction.

%-----------------------------------------------------------------------

\begin{figure*}
\begin{center}
    \subfigure[Translational (observed) flow vector $\vec{o}_t$ at pixel location $(x,y)$.]
    {
        \includegraphics[width=0.2\textwidth]{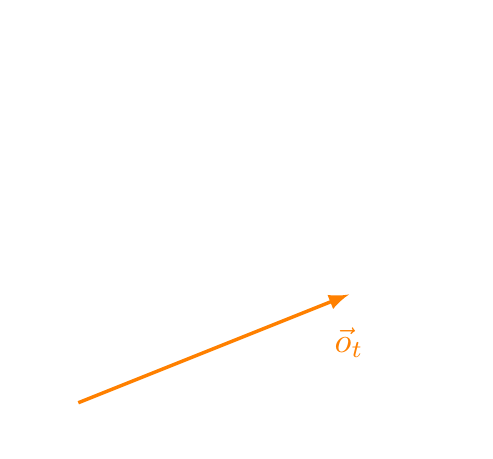}
    }\hspace{2em}
    \subfigure[Observed optical flow vector $\vec{o}_t$, is a noisy observation of the motion field vector $\vec{v}_t$: $\vec{o_t} = \vec{v}_t + \vec{n}$.]
    {
        \includegraphics[width=0.2\textwidth]{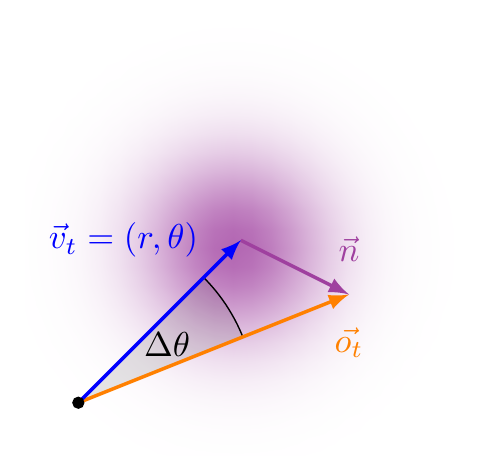}
    }\hspace{2em}
    \subfigure[To compute the flow likelihood, we integrate over the unknown motion magnitude of the motion field vector $\vec{v}_t$.]
    {
        \includegraphics[width=0.2\textwidth]{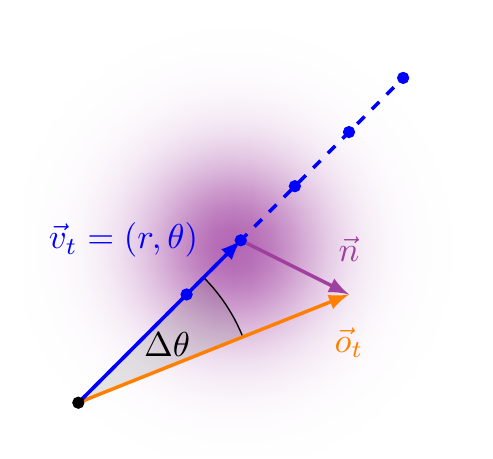}
    }\hspace{2em}
    \subfigure[Probability distribution over inverse depth.]
    {
        \includegraphics[width=0.2\textwidth]{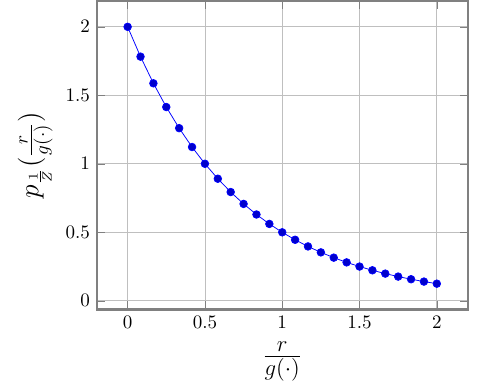}
        \label{fig:invDepth}
    }
   \caption[The flow likelihood]{{\bf Flow likelihood.} (a)-(c): computation of the probability $p(\vec{n}$) at pixel location $(x,y)$. (d): probability distribution over inverse depth. The flow likelihood is maximal, when the observed flow vector $\vec{o}_t$ and the motion field vector $\vec{v}_t$ point into the same direction with similar magnitude.}
   \label{fig:likelihood-noise}
\end{center}
\end{figure*}

\subsection{The Right Spin: Camera Motion Estimation}\label{sec:cam-estimation}
To rectify the observed optical flow field for camera rotation, we require an accurate estimate for rotation. How can we obtain a good estimate of the camera rotation and the translational motion direction that together best explain the observed motion field? Towards finding an answer to this question, we derive a novel maximum likelihood approach that aims at finding the rotation $[A,B,C]$ such that the likelihood of the resulting translational flow field is maximized. 
To this end, we derive a new flow likelihood function incorporating a model for the optical flow's noise as well as a prior distribution over the {\em inverse scene depth}.

In the following, we first introduce the new flow likelihood (Section~\ref{sec:flow-likelihood}). 
We then describe how camera motion parameters are estimated by maximizing this new likelihood function.

\subsubsection{Likelihood of the translational motion field}\label{sec:flow-likelihood}
 Let $\vec{o}_t$ be the observed translational flow vector, e.g., flow estimated with \citep{sun2018pwc}, at the pixel position $(x,y)$. Let the translational 3D motion direction of the camera $[U,V,W]$ be a unit vector. The three translational camera parameters $[U,V,W]$ and the pixel position $(x,y)$ define the direction of a motion field vector on the image plane .  As derived in~\citep{bideauCVPR2018}, the probability of observing  $\vec{o}_t$ at $(x,y)$ given a motion direction $[U,V,W]$ is given by:
\begin{align}
p(\vec{o}_t & \mid U, V, W, x, y)\nonumber\\
&=\int_0^{\infty} p(\vec{n})\;p_r(r \mid U, V, W, x, y)  \,dr, \label{eq:likelihood}
\end{align}
where $r$ denotes the magnitude of a motion field vector and $\vec{n}$ is the optical flow's noise.
This likelihood function depends on the distribution over the optical flow's noise $p(\vec{n})$ as well as the distribution over motion field magnitudes $p_r$. Figure~\ref{fig:likelihood-noise} pictures the computation of $p(\vec{n})$. Modeling the probability distribution over flow magnitudes is challenging, since those depend on the camera's translational motion direction $[U,V,W]$, the pixel location as well as the scene depth at that location. \citep{bideauCVPR2018} model $p_r$ by assuming that the motion field magnitude $r$ is independent of the flow direction $[U,V,W]$. However this often does not lead to accurate estimations, especially in the case of strong z-motion (forward motion). Here, motion field magnitudes close to the focus of expansion are near zero and the motion vectors farther away from the focus of expansion show larger magnitudes, thus the motion field magnitude is clearly dependent upon the camera's motion direction.

Next, we present a new way of modeling the distribution over motion field magnitudes $p_r$ that alleviates this problem.

\begin{comment}
\paragraph{From flow magnitudes to inverse depth.}
Given the motion field of a pure translational motion, from the perspective projection equations~\citep{bruss1983passive}, we can derive the motion field components $u_t$ and $v_t$ as:
\begin{align}
u_t&=\frac{-fU + xW}{Z}, & v_t&=\frac{-fV + yW}{Z}.
\end{align}
Then the motion field magnitude $r$ is:
\begin{align}
r&=\sqrt{u_t^2+v_t^2},\\
&=\frac{1}{Z} \cdot \sqrt{(-fU+xW)^2+(-fV+yW)^2},\\
&=\frac{1}{Z} \cdot g(f,x,y,U,V,W), \label{eqn:rgrelation}
\end{align}
where $g$ is a function that controls all aspects of the magnitude that are {\em not} related to depth.
\end{comment}

\paragraph{Distributions over flow magnitudes}
We express the motion field magnitudes in terms of inverse depth $\frac{1}Z{}$ and $g(\cdot)$, which is a function comprising all aspects of the magnitude that are {\em not} related to depth,
\begin{align}
r&=\sqrt{u_t^2+v_t^2},\nonumber\\
&=\frac{1}{Z} \cdot g(f,x,y,U,V,W). \label{eqn:rgrelation}
\end{align}
Given this reformulation of the magnitude $r$, we can determine the {\em induced distribution} over motion field magnitudes, given the distribution over inverse depths. We aim to compute $p_r(r\mid g(f,x,y,U,V,W))$ through
$p_{\frac{1}{Z}}(\frac{1}{z})$, which is the distribution over inverse depth. 
Using the relation between $r$ and $g(\cdot)$ from Eq.~\ref{eqn:rgrelation}, we can rewrite $p_r(r\mid g(\cdot))$ as follows
\begin{equation}
\label{ref:change1}
p_r(r\mid g(\cdot)) = \frac{p_{\frac{1}{Z}}(\frac{r}{g(\cdot)})}{g(\cdot)}.
\end{equation}
This is effectively just a change of units. 
Expressing the distribution over flow magnitudes in terms of the distribution over inverse depth however brings a significant advantage. This formulation effectively factors motion direction $(U,V,W)$, focal length $f$ and scene depth into the function $g(\cdot)$, and the distribution over depth can be modeled without relying on these dependencies that require making further approximations.

\paragraph{Flow likelihood}
Following prior derivations, the flow likelihood function (Eq. \ref{eq:likelihood}) can be expressed by the distribution over inverse depth, instead of flow magnitudes:
\begin{align}
p(\vec{o_t} & \mid U, V, W, x, y)\nonumber\\
&=\int_0^{\infty} p(\vec{n})\;p_r(r \mid g(\cdot))  \,dr \nonumber\\ 
&= \int_0^{\infty} p(\vec{n})\;\frac{p_{\frac{1}{Z}}\left(\frac{r}{g(\cdot)}\right)}{g(\cdot)}  \,dr. \label{eq:final-likelihood}
\end{align}
The key advantage of this is that while flow magnitudes are not independent of the motion direction, the inverse depths are, and thus the model is more realistic.

\subsubsection{Camera motion estimation via likelihood maximization}\label{sec:cam-est-trans}
\begin{figure*}
\centering
\subfigure[video frame]
{
	\includegraphics[height=0.157\textwidth]{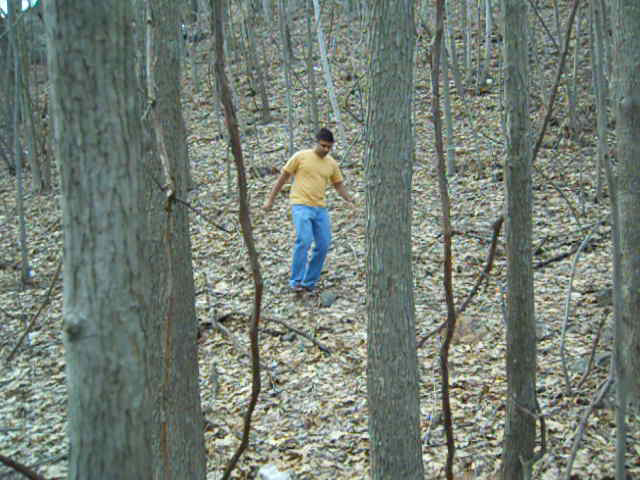}
	\includegraphics[height=0.157\textwidth]{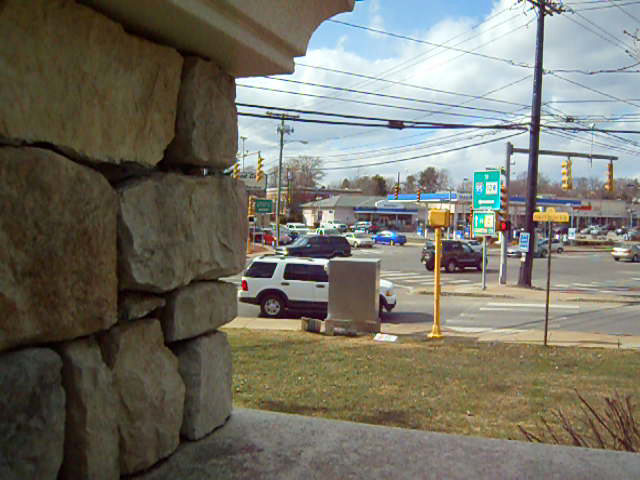}
	\includegraphics[height=0.157\textwidth]{parkour.jpg}
	\includegraphics[height=0.157\textwidth]{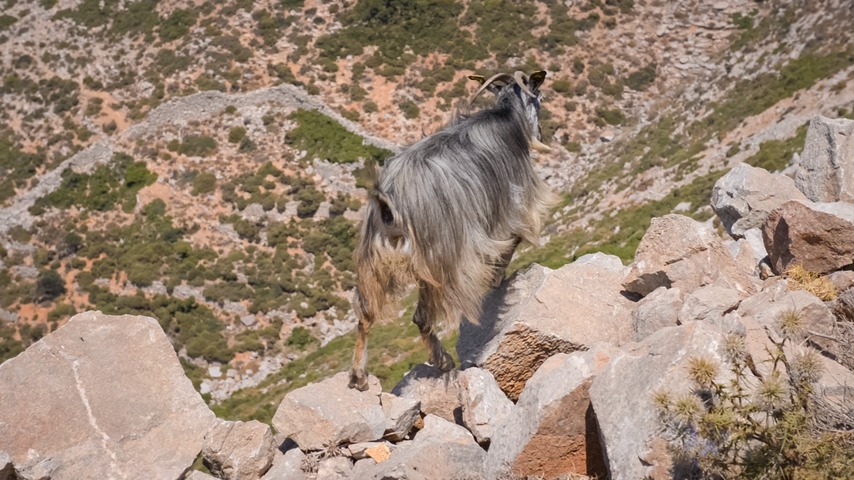}
    \label{fig:overview-flowields-frame}
}\\
\subfigure[optical flow]
{
	\includegraphics[height=0.15\textwidth]{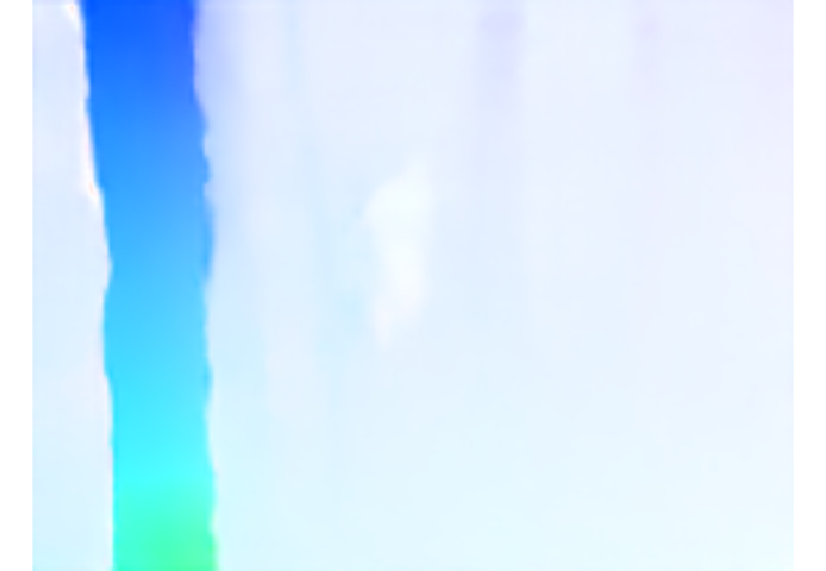}
	\includegraphics[height=0.15\textwidth]{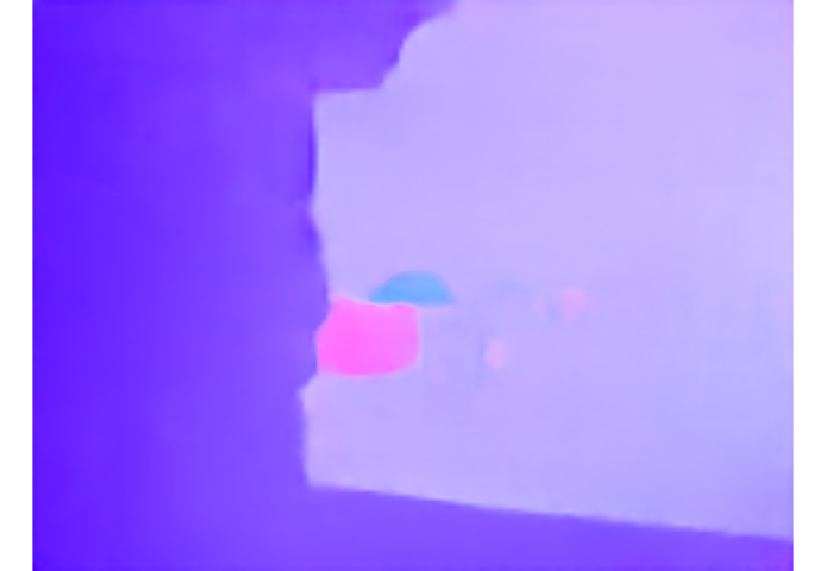}
	\includegraphics[height=0.15\textwidth]{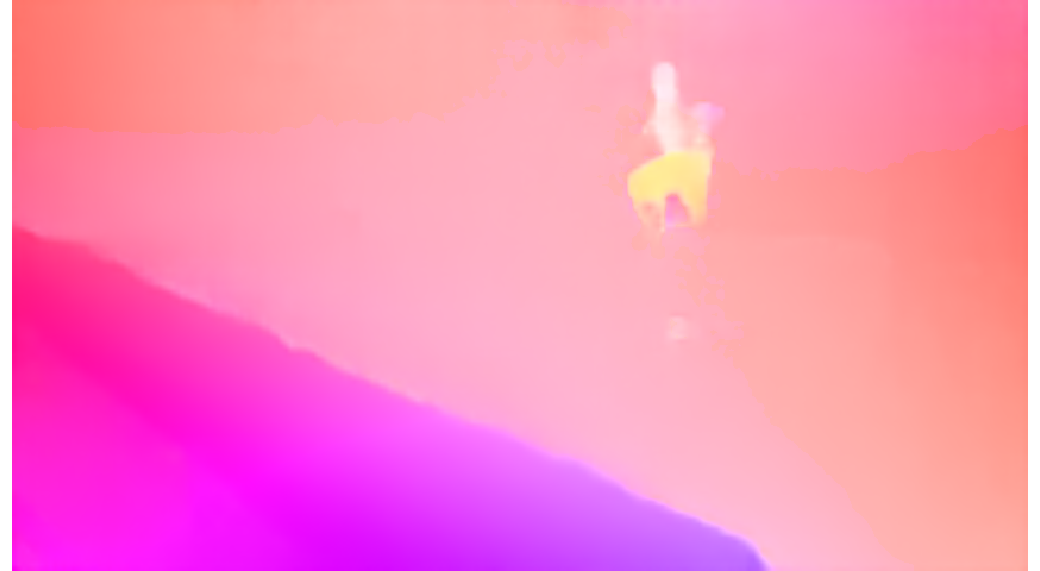}
	\includegraphics[height=0.15\textwidth]{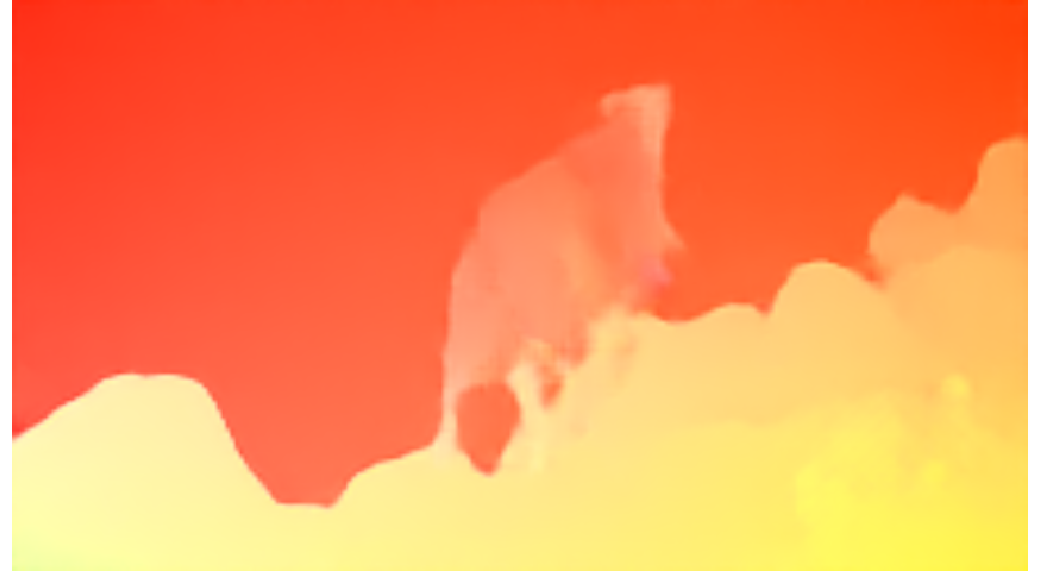}
    \label{fig:overview-flowields-flow}
}\\
\subfigure[rotation compensated optical flow]
{
    \includegraphics[height=0.15\textwidth]{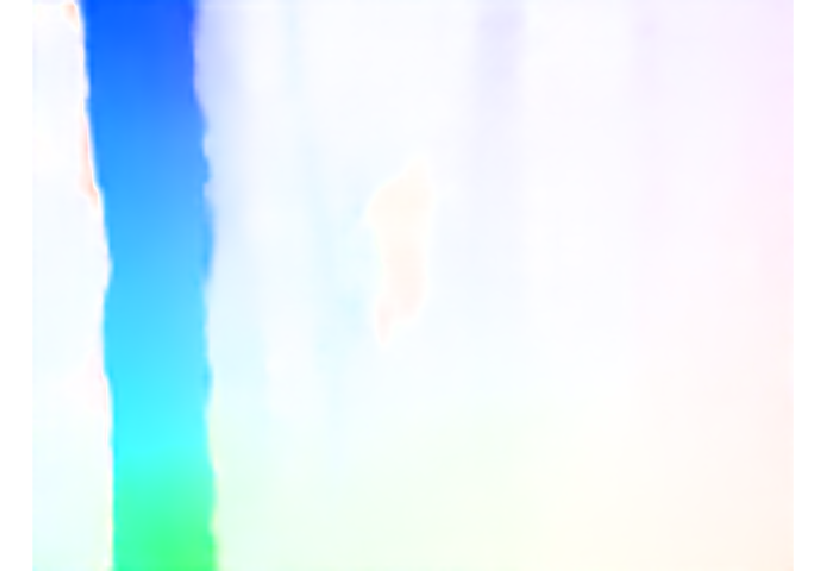}
	\includegraphics[height=0.15\textwidth]{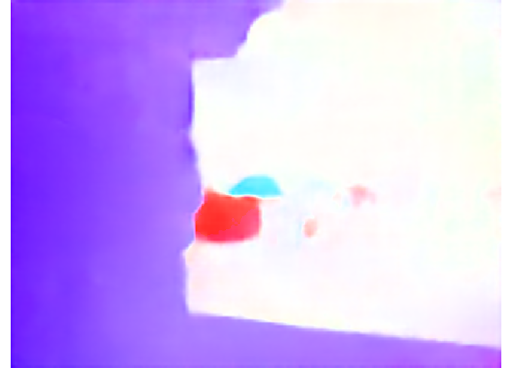}
	\includegraphics[height=0.15\textwidth]{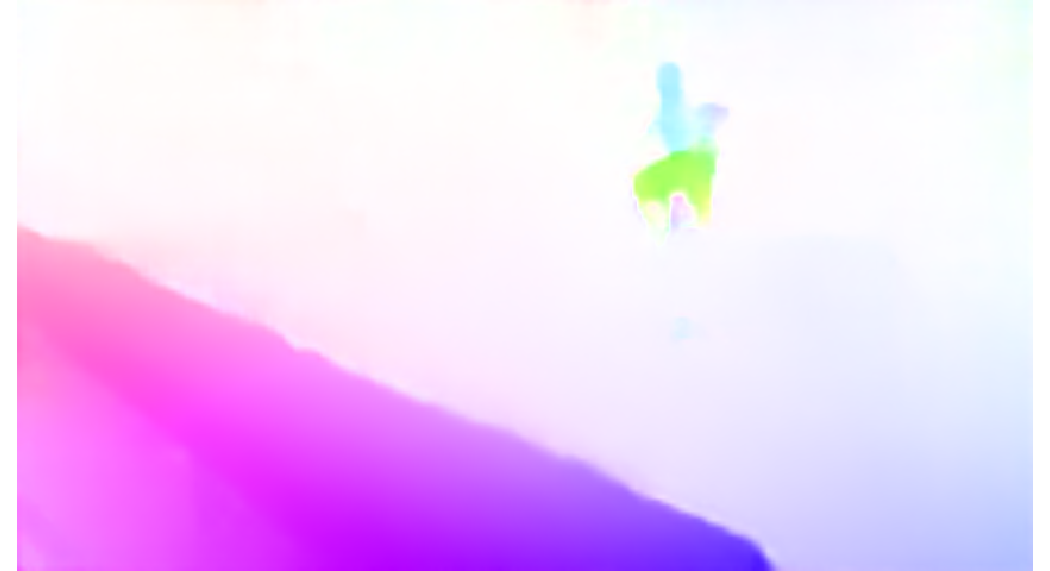}
	\includegraphics[height=0.15\textwidth]{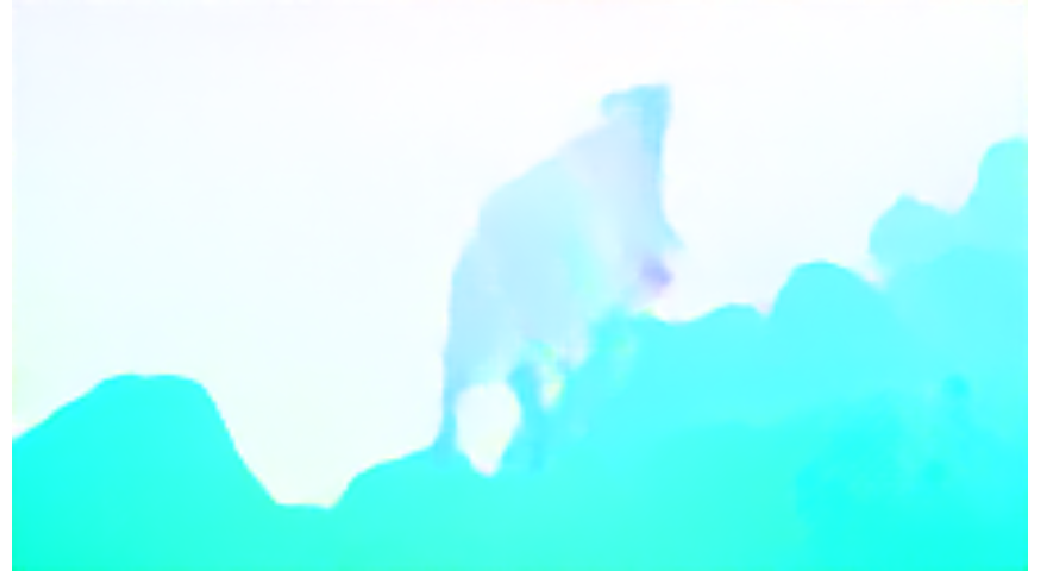}
    \label{fig:overview-flowields-trans-flow}
}\\
\subfigure[depth estimate]
{
	\includegraphics[height=0.15\textwidth]{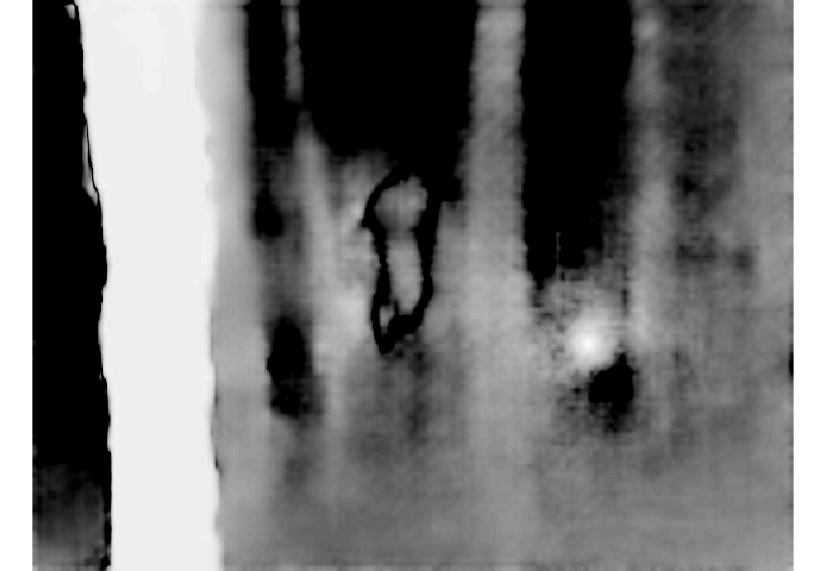}
	\includegraphics[height=0.15\textwidth]{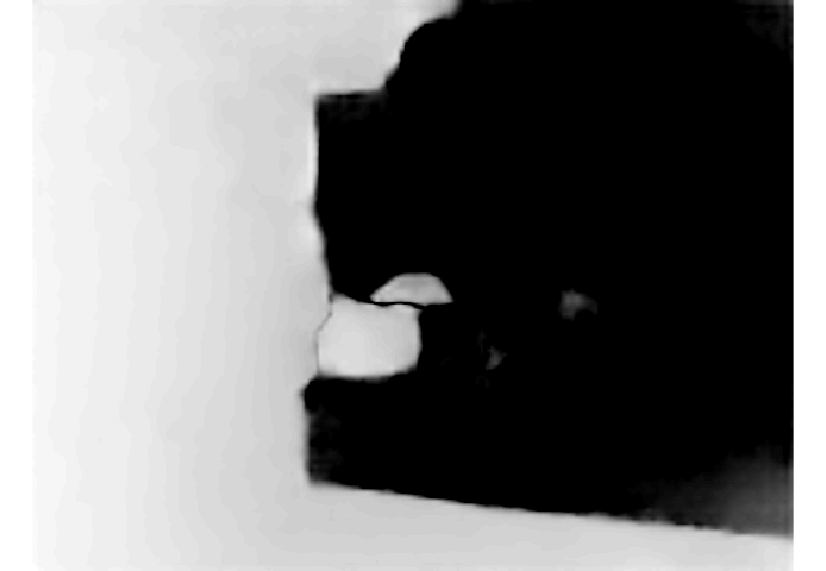}
	\includegraphics[height=0.15\textwidth]{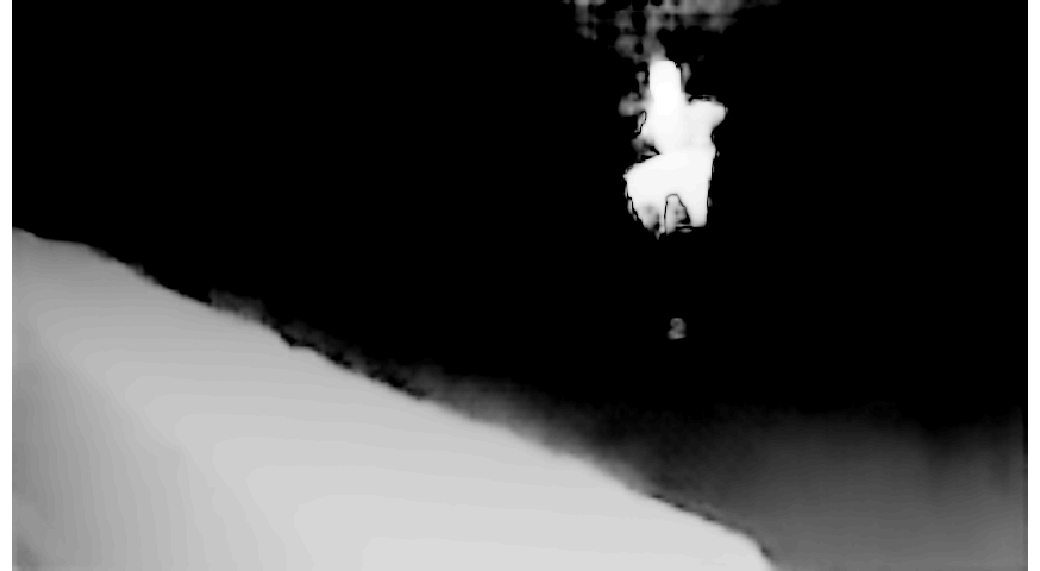}
	\includegraphics[height=0.15\textwidth]{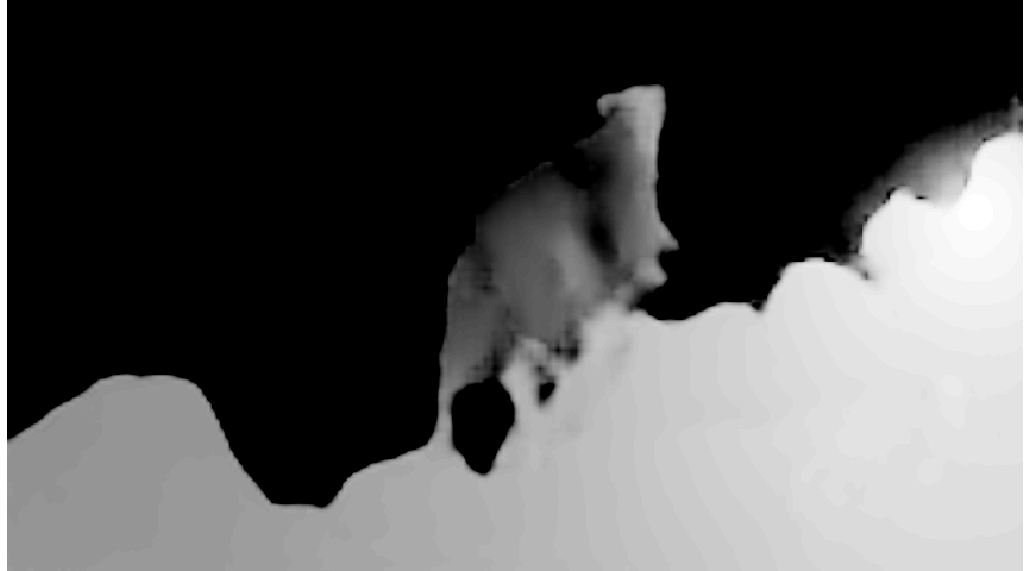}
    \label{fig:overview-flowields-depth}
}
\caption[Flow, rotation compensated flow and the relative depth estimates.]{{\bf Flow, rotation compensated flow and the relative depth estimate.} We show sample videos from the data set Complex Background (video sequences: traffic, forest) as well as two sample videos from the Davis data set (video sequence: parkour, goat). A comparison of (b) and (d) shows how motion at distant is dominated by camera rotation. After subtracting of the camera's rotation the remaining flow magnitude in these areas is very small (light color). If the flow magnitude is small the motion direction is noisy. This can be seen in (e).}
\label{fig:overview-flowields}
\end{figure*}

In Section \ref{sec:flow-likelihood}, we have derived a new likelihood function of an observed optical flow vector $\vec{o}$.
Our goal is now to find a camera rotation $(A, B, C)$ and translational camera motion direction $(U,V,W)$, such that the flow likelihood is maximal or alternatively the negative log-likelihood is minimal. 
Recall $\vec{o}_t$ is the observed translational flow vector after subtracting the flow $\vec{v}_r$ due to camera rotation:
\begin{align}
    \vec{o}_t = \vec{o} - \vec{v}_r(A,B,C).
\end{align}
Given the rotation compensated flow, we minimize the negative log-likelihood as follows:
\begin{align}
&A^*,B^*,C^*,U^*,V^*,W^*\nonumber\\
&= \operatorname*{arg\,min}_{A,B,C,U,V,W} \sum -\log(p(\vec{o_t}\mid U, V, W, x, y)).
\end{align}
Local minima are a concern when solving this optimization problem, especially in cases of noisy optical flow, inaccurate estimates of independently moving objects or complex scene geometry. To reduce this risk, we initialize the optimization using three different starting points: (1) camera rotation and translation estimate of the previous frame, (2) camera rotation estimate weighted by depth estimate of the previous frame and the translation estimate of the previous frame, and (3) camera rotation estimate weighted by depth estimate of the previous frame and the translation estimate of the previous frame in the opposite direction.
The first initialization is a good assumption if the camera motion is approximately constant. 
Initialization (2) and (3) incorporate depth information. The apparent motion of areas far away is mainly influenced by the camera's rotation and not the camera's translation (see Figure~\ref{fig:overview-flowields}), thus knowing the depth helps to correctly disentangle flow due to camera rotation and flow due to translation.

During the optimization each pixel is weighted using learned, soft object motion masks of the previous frame, that evolve over time - thus the influence of moving objects is suppressed due to a low weight. The following Section describes how object motion masks are learned while pertaining important geometric information.

\begin{figure}
\begin{center}
\includegraphics[width=0.8\columnwidth]{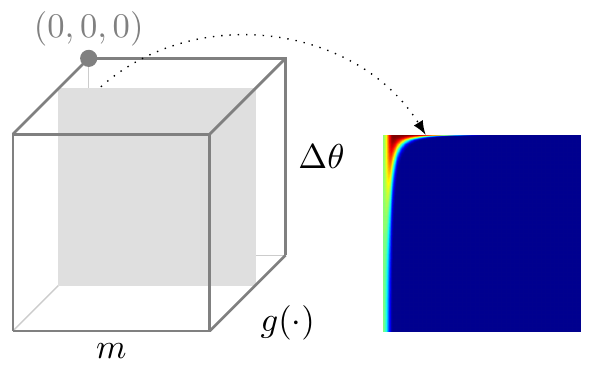}
   \caption[A 3D lookup table for the flow likelihhood.]{{\bf Lookup table picturing flow likelihood values.} Our new flow likelihood addresses the challenge of estimating the camera's motion in the presence of noisy optical flow. The color \textit{red} indicates high likelihood values, \textit{dark blue} indicates low likelihood values. The lower the angle difference $\Delta \theta$ between the vectors $\vec{o}_t$ and $\vec{v}_t$, the higher the likelihood. Note that for very small flow magnitudes $m$ the flow likelihood is almost the same regardless $\Delta \theta$. This is an important consequence of our model, indicating the unreliability of the flow direction in case of near zero magnitudes.}
   \label{fig:lookup}
\end{center}
\end{figure}

\subsection{Object Motion Segmentation}\label{sec:learning-motion}
We build our segmentation framework on an effective model for motion segmentation, that learns object motion patterns from optical flow and segments a flow field into static background and moving objects~\citep{Tokmakov17}. Yet, this model does not incorporate any geometrical concepts.
As discussed earlier optical flow fields couple information about scene geometry as well as camera motion, making the judgment whether an object is moving challenging.
By introducing a simple pre-processing step we show, that the complexity of optical flow patterns is dramatically reduced.
Different from prior work, our network processes rotation compensated flow fields (angle + magnitude) to segment independently moving objects. Learning object motion based on pre-processed flow fields appears to be an easier task to learn.
While our network architecture is similar to~\citep{Tokmakov17}, we propose important modifications to the training procedure in the following.

\subsubsection{Incorporating geometric information into training} \label{sec:training}
The network follows the classical U-net architecture and is trained on estimated translational flow fields. During training, we first estimate optical flow using~\citep{sun2018pwc} on the FlyingThings3D data set~\citep{mayer2016large}. The ground truth camera rotation is provided and subtracted from the estimated flow to obtain a rotation-compensated flow field. This flow field is input to our network. The input has a size of $h \times w \times 3$. The third dimension denotes the flow expressed in terms of angle (represented as a unit vector) and magnitude.
A representation of the flow angle as unit vector instead of angles in degree avoids segmentation discontinuities at 0 degree (or $2\pi$ respectively). The normalized flow field and the flow's magnitude are concatenated and form the input to our network. An interesting question for training a network with rotation-compensated optical flow is, whether it is worthwhile to incorporate the magnitude into the training procedure. On the one hand 
the flow magnitude can be a good indicator about the reliability of the flow angle~\citep{bideau2016s}, while on the other hand variation in larger magnitudes can be either due to variances in the scene depth or fast moving objects - thus including the magnitude might add rather misleading information. We take a closer look into this question as part of our ablation study in Section~\ref{sec:netvariants}.

\subsection{Implementation details} 

To find the camera rotation and translational motion direction that best explains the observed optical flow field, we derived a new flow likelihood function (Section~\ref{sec:cam-estimation}). Details regarding parametrization are provided in the following. 

The probability of the flow noise $p(\vec{n})$ is modeled as a multivariate normal distribution
$p(\vec{n}) \sim \mathcal{N}(\mu, \Sigma)$
and the inverse depth $p(\frac{1}{Z})$ as
an exponential distribution
$p(\frac{1}{Z}) \sim \textrm{Exp}(\lambda)$.
The noise covariance $\Sigma$ is assumed to be spherical and is measured using the ground truth flow of Sintel \citep{Butler:ECCV:2012} and the corresponding noisy estimate~\cite{sun2018pwc}.
We obtain $\Sigma=16.5\cdot10^{-5} I$, where $I$ is the identity matrix.
$\lambda$ is the rate parameter of the exponential distribution modeling the inverse depth, and is estimated using ground truth depths from Sintel.
We measured $\lambda=0.64$. The distribution over inverse depth can be seen in Figure~\ref{fig:invDepth}.

For computational efficiency the integral in Eq.~\ref{eq:final-likelihood} is approximated using a discrete sum over motion field magnitudes $r$. 
Flow likelihood values are pre-computed and stored in a lookup table for efficiency (see Figure~\ref{fig:lookup}). 

% ---------------------------------------------------------------
\section{Experiments}\label{sec:experiments}
We begin with a brief description of data sets used for training and evaluation of our motion segmentation network. 
In Section ~\ref{sec:exp-estimFlow}, we evaluate our here presented motion segmentation approach on the widely used DAVIS~\citep{perazzi2016benchmark} data set and MoCA~\citep{lamdouar2020betrayed}. Ground truth camera motion is not provided for these data sets, thus synthetic data - such as the FlyingThings3D data set~\citep{mayer2016large} and Sintel~\citep{Butler:ECCV:2012,Wulff:ECCVws:2012} - are used for ablation studies.
These studies in particular focus on the analysis of different variants of our core network and the quality as well as the effect of rotation estimation via likelihood maximization.

\textbf{\textit{DAVIS2016 (Densely Annotated VIdeo Segmentation)}}~\citep{perazzi2016benchmark} contains 50 video sequences in total with moving objects in various environments. A 30/20 training/validation split is provided. Our model is evaluated on the validation set. Ground truth segmentations of the most prominent moving object are provided for each frame. DAVIS has been widely used for general video segmentation as well as motion segmentation.

\textbf{\textit{MoCA (Moving camouflaged animals)}}~\citep{lamdouar2020betrayed} comprises a set of 141 videos depicting 67 different animals. The data set is split into three motion types describing the animals motion - {\it locomotion}, {\it static} and {\it deformation}. Following the procedure of~\citep{Lamdouar21, yang2019unsupervised} we evaluate on the {\it locomotion} split, which forms the largest part of the dataset with 88 video sequences in total. Annotations are provided in form of bounding boxes. An evaluation script is provided by the authors of MoCA.

\textbf{\textit{FT3D (FlyingThings3D)}}~\citep{mayer2016large} is a large optical flow data set, providing ground truth optical flow, RGB images, camera motion and depth. It is a synthetic data set showing random objects like chairs, tables, etc., flying in a 3D world along random trajectories. The data set is split into test and training sets. %We show experiments using ground truth optical flow and also the estimated optical flow from the RGB images.

\textbf{\textit{Sintel}} \citep{Butler:ECCV:2012,Wulff:ECCVws:2012} is the de facto benchmark for optical flow algorithms, containing 23 video sequences with 20 to 50 frames each. These short video sequences are taken from an animated movie. The scenes are realistically simulated. Synthetic videos are available with ground truth optical flow, depth, camera motion and material segmentation.

% ---------------------------------------------------------------
%\section{Learning object motion from synthetic training data}\label{sec:ECCV18}
%\input{sec-learning-object-motion-synthetic}
% ---------------------------------------------------------------

%\subsection{Evaluation: Learning object motion from ideal rotation compensated motion fields}\label{sec:exp-ECCV18}
%\input{sec-experiments-ECCV18workshop}

\subsection{Results}\label{sec:exp-estimFlow}

Our main framework consists of two steps (1) compensating the observed optical flow for camera rotation, and (2) segmenting the resulting optical flow in to static background and independently moving objects.
Experiments presented here are based on the DAVIS~\citep{perazzi2016benchmark} data set and the MoCA~\citep{lamdouar2020betrayed} data set, that each raise a slightly different aspect onto the motion segmentation problem. Details are described in the following. %In addition we provide in Section~\ref{sec:netvariants} results for camera rotation estimation via likelihood optimization and an ablation study examining different variants of our model.

\begin{table}
\small
\centering
\begin{tabular}{l | c | cccccccccc}
\hline
%\multicolumn{2}{c|}{} & \cite{Tokmakov17} & \cite{bideau2016s} &  &  \\ 
\multicolumn{1}{c|}{} & & LMP & TMM & Ours & Ours* \\
\hline
& Supervised & \ding{51} & \ding{55} & \ding{51} & \ding{51} \\
& RGB & \ding{55} & \ding{55} & \ding{55} & \ding{55} \\
& Flow & \ding{51} & \ding{51} & \ding{51} & \ding{51} \\
& Multi frame & \ding{55} & \ding{55} & \ding{55} & \ding{55} \\
\hline
\multirow{3}{*}{$\mathcal{J}$} & Mean $\uparrow$ & 58.4 &  40.1 & \bf\color{nonerikblue}59.7 & \cellcolor{yellow!70}\bf{62.5} \\ 
& Recall $\uparrow$& 67.3 &  34.3 & \bf\color{nonerikblue}{69.6} & \cellcolor{yellow!70}\bf{73.8} \\
& Decay $\downarrow$ & ~5.6 & 15.2 & \bf\color{nonerikblue}{~4.3} & \cellcolor{yellow!70}\bf{~3.8} \\
\hline
\multirow{3}{*}{$\mathcal{F}$} & Mean $\uparrow$ & 58.4 & 39.6 & \bf\color{nonerikblue}{59.5} & \cellcolor{yellow!70}\bf{61.1} \\
& Recall $\uparrow$ & 66.0 & 15.4 & \bf\color{nonerikblue}{66.4} & \cellcolor{yellow!70}\bf{69.9} \\
& Decay $\downarrow$& ~7.9 & 12.7 & \cellcolor{yellow!70}\bf{~5.4} & \bf\color{nonerikblue}{~5.6} \\
\hline
$\mathcal{T}$ & Mean $\downarrow$ & 87.8 & \cellcolor{yellow!70}\bf 51.3 & \bf\color{nonerikblue}74.5 & 83.4\\
\hline
\end{tabular}
\caption[Motion segmentation: Comparison to other approaches using only motion cues on DAVIS.]{Motion segmentation: Comparison to other approaches using only motion cues on DAVIS {\it(train-val)}, i.e., without any appearance. Ours refers to the variant of our model using only motion cues and no appearance terms and Ours* denotes a motion-only upper bound, which uses ground truth segmentation for camera motion estimation. Best viewed in color (\colorbox{yellow!70}{\bf1st-best}, {\bf\color{nonerikblue} 2nd-best}).}
\label{tbl:davis_ablation}
\end{table}

\paragraph{DAVIS: Optical flow only} We compare our motion segmentation network with other methods that use optical flow as the only cue for segmentation. Table~\ref{tbl:davis_ablation} shows these results on DAVIS. 
LMP~\citep{Tokmakov17} is a learning based approach trained on ground truth optical flow of FlyingThings3D. This approach relies on a simliar network architecture, but does not incorporate an explicit model for modeling geometrical concepts, e.g. the scene geometry and camera motion. TMM~\citep{bideau2016s}, on the contrary, compensates flow for camera rotation and attempts to segment a video by assigning translational motion models to different image regions in a probabilistic fashion. The exclusive usage of translational motion models however quickly leads to oversegmentations and fails to capture more complex motion patterns. While combining geometrical concepts such as perspective projection together with learned motion patterns, our approach improves over both these motion segmentation methods. The segmentation performance is measured using the $\mathcal{J}$-Mean score. We achieve an $\mathcal{J}$-Mean score of 59.7. The next best performing method is LMP resulting in an $\mathcal{J}$-Mean score of 58.4.  We compute an upper bound for our method (Ours* in Table~\ref{tbl:davis_ablation}) by masking out independently moving objects, with ground truth segments, for our camera motion estimation procedure. This masking procedure eliminates errors of our camera motion estimation due to `outliers' in optical flow, such as moving objects. %A similar result can be achieved with classical approaches such as RANSAC.

\begin{table*}
\centering
\begin{tabular}{l | c | ccc|ccccccccc}
\hline
%\multicolumn{2}{c|}{} &  \cite{Lamdouar21} & \cite{yang2021self} & \cite{yang2019unsupervised} & \cite{lamdouar2020betrayed} & \cite{Lu_2019_CVPR} & \cite{zhou2020matnet} &  &  \\
\multicolumn{1}{c|}{} & & SegI & MG & CIS & COD & COSNet & MATNet & Ours & Ours+Temp\\
\hline
\multirow{3}{*}{} & Supervised &  \ding{55} & \ding{55} &\ding{55} & \ding{51} & \ding{51} & \ding{51} & \ding{51} & \ding{51}\\
& RGB & \ding{55} & \ding{55} & \ding{51} & \ding{51} & \ding{51} & \ding{51} & \ding{55} & \ding{55}\\
& Flow &  \ding{51} & \ding{51} & \ding{51} & \ding{51} & \ding{55} & \ding{51} & \ding{51} & \ding{51}\\
& Multi-frame &  \ding{51} & \ding{51} & \ding{51} & \ding{55} & \ding{55} & \ding{51} & \ding{55} & \ding{51}\\
\hline
$\mathcal{J}$ & Mean $\downarrow$ & {\bf68.6} & 63.4 & 49.4 & 44.9 & 50.7 & \bf\color{nonerikblue}64.2 & 58.3 & \colorbox{yellow!70}{\bf65.8}\\
\hline
\multirow{6}{*}{\rotatebox[origin=c]{90}{Success Rate}} & $\tau = 0.5$ &  {\bf 77.2} & 74.2 & 55.6 & 41.4 & 58.8 & \bf\color{nonerikblue}71.2 & 64.5 &  \colorbox{yellow!70}{\bf72.7}\\
& $\tau = 0.6$ &  {\bf71.7} & 65.4 & 33.0 & 53.4 & \colorbox{yellow!70}{\bf67.0} & 46.3 & 58.0 & \bf\color{nonerikblue}65.2\\
& $\tau = 0.7$ & {\bf 62.3} & 52.4 & 32.9 & 23.5 & 45.7 & \colorbox{yellow!70}{\bf59.9} & 49.8 & \bf\color{nonerikblue}53.8\\
& $\tau = 0.8$ & {\bf 46.4} & 35.1 & 17.6 & 14.0 & 33.7 & \colorbox{yellow!70}{\bf49.2} & 36.2 & \bf\color{nonerikblue}38.9\\
& $\tau = 0.9$ & {\bf 25.5} & 14.7 & 3.0 & 5.9 & 16.7 & \colorbox{yellow!70}{\bf24.6} & 16.6 & \bf\color{nonerikblue}17.3\\
& $SR_{mean}$ & {\bf 56.6} & 48.4 & 31.1 & 23.6 & 41.7 & \colorbox{yellow!70}{\bf54.4} & 45.0 & \bf\color{nonerikblue}49.6\\
\hline
\end{tabular}
\caption[Motion segmentation: Comparison to state-of-the-art motion segmentation methods on MoCA.]{Motion segmentation: Comparison to state-of-the-art motion segmentation methods on MoCA. Methods we compare against from left to right:~\citep{Lamdouar21,yang2021self,yang2019unsupervised,lamdouar2020betrayed,Lu_2019_CVPR,zhou2020matnet}. {\bf Bold} indicates best among all methods, while \colorbox{yellow!70}{\bf1st-best} and {\bf\color{nonerikblue} 2nd-best} represent the best and second best within the supervised methods. Best viewed in color.}
\label{tbl:MoCA}
\end{table*}

\paragraph{MoCA: Optical flow only} Data sets like MoCA focus in particular on the segmentation of objects that can only be robustly recognized based on their unique motion. Where as most data sets for moving object segmentation combine several cues (motion and appearance) that are helpful for recognizing moving objects, this data set highlights the relevance of motion. Thus MoCA allows to evaluate the strengths of motion models in isolation.
It is not surprising that appearance cues are rather weak in cases of camouflage, therefore methods based on RGB frames only (e.g. COSNet~\citep{Lu_2019_CVPR}) show a weak performance in these settings (see Table~\ref{tbl:MoCA}). On more general data set like the DAVIS, those methods show a superior performance among all other methods and achieve a similar segmentation quality as MATNet~\citep{zhou2020matnet} (Table~\ref{tbl:davis_numerical_results}).

Our approach taking a single optical flow frame (compensated for camera rotation) as input, performs comparable to other supervised approaches. A simple post-processing step - convolution with a 3D Gaussian filter and frame-wise application of a dense CRF, eliminates temporal instabilities (Ours+Temp in Table~\ref{tbl:MoCA}). Among all methods SegI~\citep{Lamdouar21} shows best results on MoCA, on DAVIS their performance falls rather short due to their lack of a strong appearance model. SegI combines multiple ConvNets where each of them encode a flow frame together with a transformer network without taking RGB frames into consideration. The model is trained on synthetically generated data, thus can be considered as unsupervised. In contrast, our approach was trained using rotation compensated flow frames estimated from the synthetic dataset {\it FlyingThings3D}.

\begin{table*}
\small
\centering
\begin{tabular}{l | c | ccc|cccc|cccc}
\hline
%\multicolumn{2}{c|}{} & \cite{Cheng_ICCV_2017} & \cite{Lu_2019_CVPR} & \cite{zhou2020matnet} & \cite{Tokmakov17} & \cite{fusionseg} & \cite{Tokmakov17ICCV} &  & \cite{yang2019unsupervised} & \cite{koh2017primary} & \cite{Lamdouar21} & \cite{yang2021self}\\
\multicolumn{1}{c|}{} & & SFL & COSNet & MATNet & LMP+App & FSEG & LVO & Ours+App & CIS & ARP & SegI \\
\hline
& Supervised & \ding{68} & \ding{68} & \ding{68} & \ding{51} & \ding{51} & \ding{51} & \ding{51} & \ding{55} & \ding{55} & \ding{55} \\
& RGB & \ding{51} & \ding{51} & \ding{51} & \ding{51} & \ding{51} & \ding{51} & \ding{51} & \ding{51} & \ding{51} & \ding{55} \\
& Flow & \ding{51} & \ding{55} & \ding{51} & \ding{51} & \ding{51} & \ding{51} & \ding{51} & \ding{51} & \ding{51} & \ding{51} \\
& Multi-frame & \ding{55} & \ding{55} & \ding{51} & \ding{55} & \ding{55} & \ding{51} & \ding{51} & \ding{51} & \ding{51} & \ding{51} \\
\hline
\multirow{3}{*}{$\mathcal{J}$} & Mean $\uparrow$ & 67.4 & 80.5 & \bf82.4 & 70.0 & 70.7 & \bf\color{nonerikblue}72.2 & \colorbox{yellow!70}{\bf73.5} & 71.5 & 76.2 & 67.8 \\
& Recall $\uparrow$ & 81.4 & 94.0 & \bf94.5 & \bf\color{nonerikblue}85.0 & 83.5 & 82.4 & \colorbox{yellow!70}{\bf85.5} & 86.5 & 91.1 & - \\
& Decay $\downarrow$ & 6.2 & \bf0.0 & 5.5 & 1.3 & 1.5 & \colorbox{yellow!70}{\bf0.1} & \bf\color{nonerikblue}1.2 & 9.5 & 7.0 & - \\
\hline
\multirow{3}{*}{$\mathcal{F}$} & Mean $\uparrow$ & 66.7 & 79.4 & \bf80.7 & 65.9 & 65.3 & \bf\color{nonerikblue}67.5 & \colorbox{yellow!70}{\bf68.9} & 70.5 & 70.6 & - \\
& Recall $\uparrow$ & 77.1 & \bf90.4 & 90.2 & \bf\color{nonerikblue}79.2 & 73.8 & 75.4 & \colorbox{yellow!70}{\bf79.6} & 83.5 & 83.5 & - \\
& Decay $\downarrow$ & 5.1 & \bf0.0 & 4.5 & 2.5 & \bf\color{nonerikblue}1.8 & 2.7 & \colorbox{yellow!70}{\bf1.4} & 7.0 & 7.9 & - \\
\hline
\end{tabular}
\caption[Motion segmentation: Comparison to state-of-the-art motion segmentation methods on DAVIS.]{Motion segmentation: Comparison to state-of-the-art motion segmentation methods on DAVIS2016. We group approaches according their training strategy: supervised and trained on the DAVIS training split (\ding{68}), supervised and trained on other segmentation data sets (\ding{51}) and unsupervised methods (\ding{55}). Methods we compare against from left to right:~\citep{Cheng_ICCV_2017,Lu_2019_CVPR,zhou2020matnet,Tokmakov17,fusionseg,Tokmakov17ICCV,yang2019unsupervised,koh2017primary,Lamdouar21,yang2021self}. {\bf Bold} indicates best among all methods, while \colorbox{yellow!70}{\bf1st-best} and {\bf\color{nonerikblue} 2nd-best} represent the best and second best within the supervised methods. Best viewed in color.}
\label{tbl:davis_numerical_results}
\end{table*}

\paragraph{DAVIS: Optical flow + Appearance} Where our main contribution lays in a novel approach to learn to segment moving object based on optical flow only, we incorporate here appearance information similar to LVO~\citep{Tokmakov17} and compare to segmentation approaches that consider both - appearance as well as motion information (Table~\ref{tbl:davis_numerical_results}).
Within the group of supervised approaches our approaches shows best performance in terms of mean/recall $\mathcal{J}$ and $\mathcal{F}$. Where as ours and LVO integrate appearance cues in a similar manner, these approaches differ in the way how object motion cues are learned. LVO learns object motion patterns directly from optical flow, where as we first disentangle camera rotation and translation before segmenting independently objects. Ablation studies analyze the usefulness of this disentanglement in further detail. Within the unsupervised approaches ARP~\citep{koh2017primary}, which is a non learning based approach, reaches highest performance. Due to multiple iterations over the entire video this approach is computationally expensive as mentioned in~\citep{yang2021self}.
Among all methods MATNet reaches highest accuracy in terms of mean $\mathcal{J}$ and $\mathcal{F}$. One reason might lay in their training strategy, which makes use of the DAVIS training set (indicated with \ding{68}).

\begin{figure*}
\begin{center}
\includegraphics[width=1.0\textwidth]{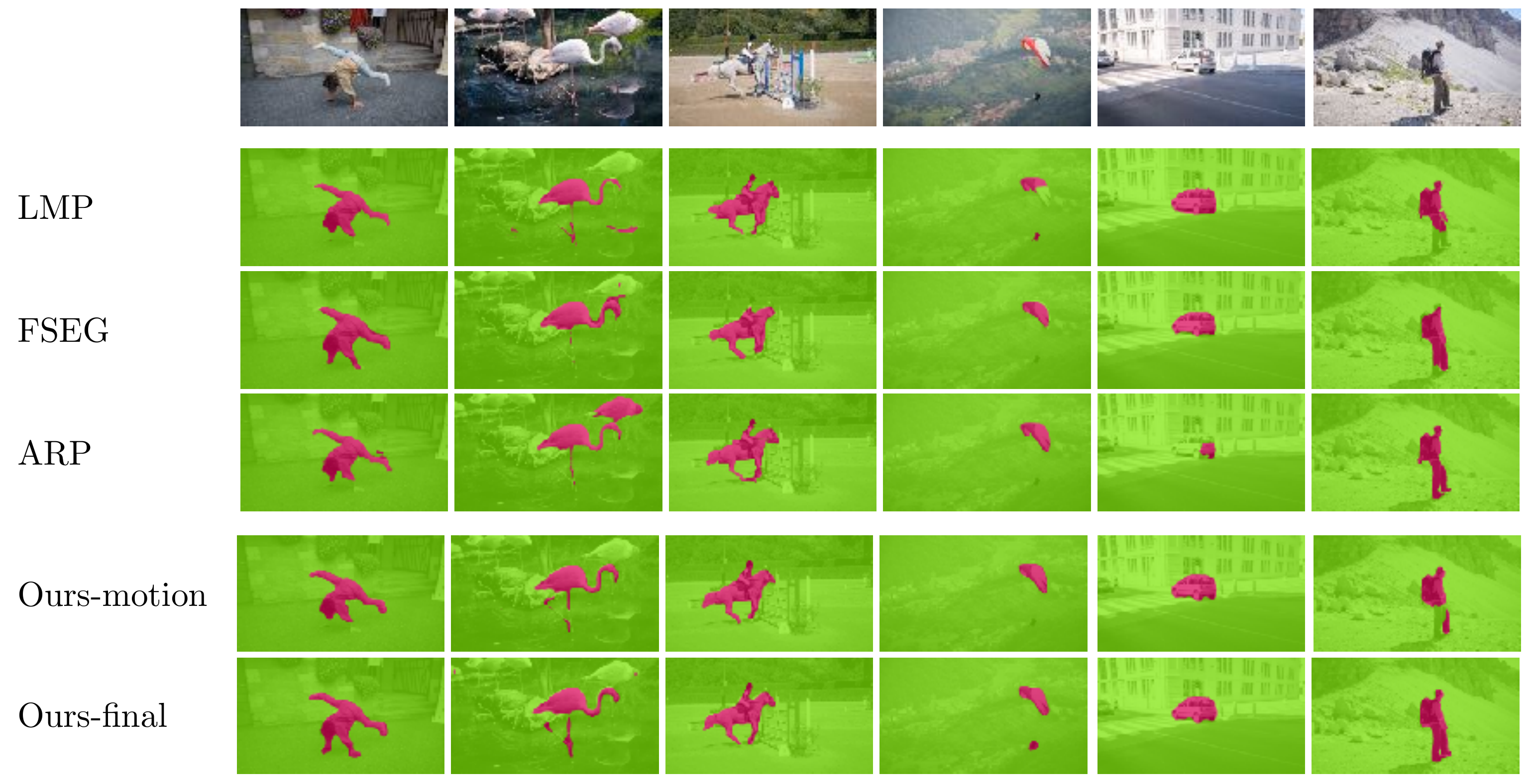}
   \caption[Qualitative segmentation results.]{{\bf Qualitative segmentation results.} Qualitative segmentation results on the DAVIS data set, showing a comparison with three other best performing methods. Ours-final denotes our complete method and Ours the variant based on motion cues alone.}
   \label{fig:results-comparison}
\end{center}
\end{figure*}

A qualitative comparison with the best performing methods is shown in Figure~\ref{fig:results-comparison}. Our results based on optical flow only and based on optical flow in combination with appearance are shown in the last two rows of this figure. These two rows in particular highlight the complementarity of motion and appearance cues. We miss the hiker's foot when relying on motion alone (Ours), since it is not moving. However, while integrating motion with appearance, we segment the entire object accurately.
ARP, the strongest method among unsupervised approaches, relies on segmenting the primary object(s) in a video and and comes with a noticable bias towards the object's appearance. In many cases such a strong appearance model is advantageous, however can lead to erroneous segmentations in other cases. For example, it only segments a part of the car (Figure~\ref{fig:results-comparison}: 2nd column from the right), which moves from the darker (shadow) area to the brighter (sunny) region.), as it matches the primary object in appearance. Our method that extracts geometrical information from optical flow and integrates learned objectness cues is capable of overcoming these types of failure cases.

\subsection{Ablation study}
\label{sec:netvariants}

\begin{figure}
\begin{center}
\includegraphics[width=0.95\columnwidth]{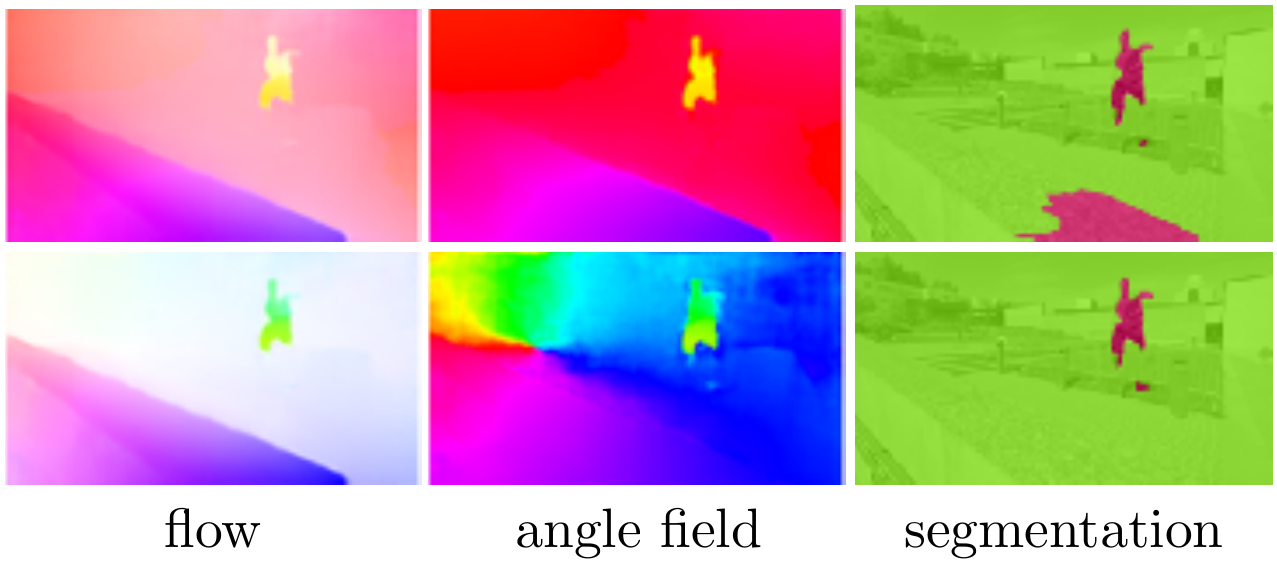}
   \caption[Comparison of motion segmentation results based on the original and the rotation-compensated flow field.]{{\bf Ablation study: Comparison of motion segmentation results based on the original and the rotation-compensated flow field.} Top row: motion segmentation with the original flow field that includes camera rotation, translation and object motion. Bottom row: motion segmentation based on \textit{rotation-compensated flow field}. Note that the angle field (middle) of the rotation-compensated flow is entirely depth independent. The angle field is fully determined by the translational camera motion and object motion. In this example one can observe a clear z-motion of the camera, which is shown by the rainbow pattern. The angle field of the original flow containing both camera rotation and translation is depth dependent (top row, middle image). This angle field clearly shows discontinuities in angle at the wall, which is due to significant changes in depth and {not} because of independent object motion.}
   \label{fig:example-parkour}
\end{center}
\end{figure}

\paragraph{Network variants} We trained four variants of our motion segmentation network, with: (1) ground truth optical flow, (2) the ground truth flow after removing ground truth camera rotation, i.e., with rotation compensated-flow fields, (3) estimated optical flow field using PWC-Net~\citep{sun2018pwc}, and (4) estimated ground truth flow compensated with ground truth camera rotation, i.e., estimated rotation compensated-flow field.
Table~\ref{table:different-trainingData} shows the analysis with these four variants. Training and testing with ground truth optical flow (original: gt flow or compensated: gt t-flow) is significantly better than using estimated optical flow. Segmentation accuracy is about 20\% higher on the FT3D test set for ground truth, compared to estimated optical flow. Training with rotation-compensated optical flow consistently leads to improved quality of the final segmentation, e.g., 90.68\% vs. 93.23\%, which supports the idea behind our method. Learning can be significantly simplified, if we are able to efficiently incorporate knowledge about physical concepts into the process of moving object segmentation. 
A direct comparison in terms of segmentation quality between using the original optical flow as input instead of the rotation-compensated optical flow is shown in Figure~\ref{fig:example-parkour}.

\begin{table}[b]
\begin{center}
\small
 \centering
\begin{tabular}
%{p{0.1\columnwidth}c{0.03\columnwidth}c{0.03\columnwidth}c{0.03\columnwidth}c{0.03\columnwidth}} 
{llc}
\toprule 
trained with... & tested with... & {angle+magnitude} \\
\midrule 
gt flow & gt flow &  90.68 \\ \addlinespace[0.02cm]
gt t-flow & gt t-flow &  ~\bf93.23 \\ \addlinespace[0.02cm]
%gt FT3D & PWC-Net FT3D &  74.40 \\ \addlinespace[0.02cm]
%gt transFT3D & PWC-Net transFT3D & 24.44 \\ \addlinespace[0.02cm]
PWC flow & PWC flow &  77.18 \\ \addlinespace[0.02cm]
PWC t-flow & PWC t-flow & \bf78.69  \\ \addlinespace[0.02cm]
\bottomrule
\end{tabular}
\end{center}
\caption[Ablation study: Network variants.]{{Ablation study: Network variants.} We trained four networks using flow angle and magnitude with: the provided ground truth optical flow of FT3D~\citep{mayer2016large} (gt flow), ground truth optical flow after subtracting ground truth camera rotation (gt t-flow), estimated optical flow using~\citep{sun2018pwc} (PWC flow), and estimated optical flow after subtracting ground truth camera rotation (PWC t-flow). Segmentation accuracy is measured on the FT3D test set with intersection over union (IoU) scores.}
\label{table:different-trainingData}
\end{table}

\paragraph{Training on flow angle only versus angle+magnitude} As discussed in Section~\ref{sec:motionfield}, rotation-compensated flow comprises all the information about independent object motion and the scene structure (depth). In this context, two interesting questions to tackle are: \textit{how well can one extract information about independent object motion from the angle alone}, and \textit{does including the flow magnitude (training the network on the full optical flow) improve motion segmentation?}. We show this analysis in Table~\ref{table:angleVangle+magnitude}, with further variants of our network. Using angle and magnitude together (angle+magn in the table) leads to the best performance. However, note that we achieve reasonable segmentation quality even when using the flow angle alone. The network trained on ground truth optical flow adapts very poorly to estimated optical flow, with the segmentation accuracy dropping from 93.23\% to 24.44\% for the angle+magn variant.

\begin{table}
\begin{center}
\small
 \centering
\begin{tabular}
%{p{0.1\columnwidth}c{0.03\columnwidth}c{0.03\columnwidth}c{0.03\columnwidth}c{0.03\columnwidth}} 
{ll*{2}{c}}
\toprule 
trained with... & tested with... & {angle} & {angle+magn} \\
\midrule 
gt t-flow & gt t-flow  & 77.47 &  93.23 \\ \addlinespace[0.02cm]
gt t-flow & PWC t-flow  & 24.06 & 24.44 \\ \addlinespace[0.02cm]
PWC t-flow & PWC t-flow  & 77.79 & \bf 78.69  \\ \addlinespace[0.02cm]
\bottomrule
\end{tabular}
\end{center}
\caption[Ablation study: Training with angle vs angle and magnitude.]{{Ablation study: Training with angle vs angle and magnitude.} We trained four variants of our segmentation network with: (1)~angle of the rotation-compensated flow of FT3D, (2)~angle and magnitude of the rotation-compensated flow of FT3D (angle+magn), (3)~angle of the estimated rotation-compensated flow, and (4)~angle and magnitude of the estimated rotation-compensated flow. We show consistently better performance by including magnitude. The performance is the worst when the network is trained on the angle of the rotation-compensated ground truth flow. Here, the noise in angle leads to a very significant drop on estimated optical flow data. Segmentation accuracy is measured on the FT3D test set with intersection over union (IoU).}
\label{table:angleVangle+magnitude}
\end{table}

\paragraph{Rotation estimation via likelihood maximization}

We show results on the Sintel data set (Table~\ref{tbl:sintel_ablation}), and compare our new likelihood optimization procedure with~\citep{bideau2016s}. The ground truth optical flow and focal length is provided, so an accurate estimate of the camera's rotation is possible.
Our camera rotation estimation based on maximizing the flow likelihood shows consistently better results on the Sintel data set. More importantly, the performance gap gets significant when using estimated flow as input for camera motion estimation. Since our proposed optimization approach incorporates an explicit noise model, it is significantly more robust to noisy flow data.

\begin{table}
\small
\centering
\begin{tabular}{l  c  c}
\toprule
 & TMM & Ours \\
 \midrule
gt-flow & 0.08 / 0.22 / 0.02 & 0.03 / 0.06 / 0.01 \\
PWC-flow & 0.13 / 0.34 / 0.04 & 0.05 / 0.11 / 0.03 \\
\bottomrule
\end{tabular}
\caption[Ablation study: Camera rotation estimation.]{{Ablation study: Camera rotation estimation.} Avg.\ yaw/pitch/roll error in degrees between two consecutive frames. 
\textit{gt-flow, PWC-flow:} To evaluate rotation estimation we used ground-truth segmentation masks to weight the optim.\ loss. Thus, errors in the segmentation procedure are not propagated throughout the video.
}
\label{tbl:sintel_ablation}
\end{table}

%\bmhead{Acknowledgments}

%Acknowledgments are not compulsory. Where included they should be brief. Grant or contribution numbers may be acknowledged.

%Please refer to Journal-level guidance for any specific requirements.

\section*{Declarations}

%Some journals require declarations to be submitted in a standardised format. Please check the Instructions for Authors of the journal to which you are submitting to see if you need to complete this section. If yes, your manuscript must contain the following sections under the heading `Declarations':
\subsection*{Funding}
This work was supported in part by the ANR grant AVENUE (ANR-18-CE23-0011) and Deutsche Forschungsgemeinschaft (DFG, German Research Foundation) under Germany’s Excellence Strategy – EXC 2002/1 ``Science of Intelligence" – project number 390523135.

\subsection*{Conflicts of interest/competing interests}
The authors have no conflicts of interest to declare that are relevant to the content of this article.

%%\bibliography{sn-bibliography}% common bib file
\bibliography{IJCV}% common bib file


\end{document}